\documentclass[12pt]{article}
\usepackage{titlesec}
\usepackage[T1]{fontenc}
\usepackage[utf8]{inputenc}
\usepackage{authblk}

    \usepackage{amsfonts}
    \usepackage{amssymb}
    \usepackage{amsbsy}
    \usepackage{amsmath}
    \usepackage{amsthm}
    \usepackage{bbm}
    \usepackage{bm}
    \usepackage{mathtools}
    \newcommand\independent{\protect\mathpalette{\protect\independenT}{\perp}}
\def\independenT#1#2{\mathrel{\rlap{$#1#2$}\mkern2mu{#1#2}}}

    \newtheorem{prop}{Proposition}
    \newtheorem{assum}{Assumption}
\newcounter{subassumption}[assum]
	\renewcommand{\thesubassumption}{(\textit{\roman{subassumption}})}
	\makeatletter
	\renewcommand{\p@subassumption}{\theassum}
	\makeatother
	\newcommand{\subasu}{
  \refstepcounter{subassumption}%
  \thesubassumption~\ignorespaces}

	\newtheorem*{obs*}{Observation}

    \usepackage[flushleft,online,para]{threeparttable}
	\usepackage{graphics}
\usepackage[vlined,linesnumbered,ruled,resetcount]{algorithm2e}
\SetKwInOut{Initialization}{Initialization}

    \usepackage{subfig}
    \graphicspath{{Graphs/}}
    \usepackage{multirow}

    \usepackage{tikz}
    \usetikzlibrary{arrows}
    \usetikzlibrary{decorations.pathreplacing}
    \usepackage{pgfplots}
\usepackage{indentfirst}             
\usepackage{floatrow}
    \usepackage{tabularx}
        \newcolumntype{Z}{>{\centering\arraybackslash}X}
        \newcolumntype{L}{>{\raggedright\arraybackslash}X}
    \usepackage{dcolumn}
        \newcolumntype{d}[1]{D{.}{.}{#1}}
    \usepackage{rotating}                     
    \usepackage{lscape}
    \usepackage{pdflscape}   
\usepackage{authblk}

	\usepackage{csquotes}
	\usepackage[round]{natbib}
	\usepackage{url}
	\let\OLDthebibliography\thebibliography
\renewcommand\thebibliography[1]{
  \OLDthebibliography{#1}
  \setlength{\parskip}{0pt}
  \setlength{\itemsep}{0pt plus 0.3ex}
}

    \usepackage{xcolor}
        \definecolor{darkblue}{rgb}{0,0,0.4}
    \usepackage{hyperref}
        \hypersetup{
            colorlinks = true,
            linkcolor = darkblue,
            citecolor = darkblue,
            pdfborder = 0 0 0,
            pdfdisplaydoctitle = true,
            pdfhighlight = /N,
            pdfpagelayout = OneColumn,
            pdfpagemode = UseNone,
            pdfstartview = {FitH},
            pdfauthor = {{AB, CF \& JR}},
            pdftitle = {{bits}},
            pdfsubject = {{}}
        }

    \usepackage[textsize=footnotesize, colorinlistoftodos, textwidth=4cm, obeyDraft]{todonotes}
    \usepackage{geometry}
    \usepackage{setspace}
    \usepackage[bottom, multiple,flushmargin]{footmisc}  
    \usepackage{verbatim}
    \usepackage[normalem]{ulem}     
    \usepackage{mathpazo}
    \titlespacing*{\section}
{0pt}{0.1\baselineskip}{0.1\baselineskip}
    \titlespacing*{\subsection}
{0pt}{0.1\baselineskip}{0.1\baselineskip}
    \titlespacing*{\subsubsection}
{0pt}{0.1\baselineskip}{0.1\baselineskip}
    \titlespacing*{\paragraph}
{0pt}{0.1\baselineskip}{0.1\baselineskip}
    \titlespacing*{\assum}
{0pt}{0.1\baselineskip}{0.1\baselineskip}
\titlespacing*{\table}
{0pt}{0.1\baselineskip}{0.1\baselineskip}
\titlespacing*{\align}
{0pt}{0.1\baselineskip}{0.1\baselineskip}

    \usepackage[toc,page]{appendix}
	\usepackage{lipsum}

\begin{document}

\begin{spacing}{0}

\title{Online Causal Inference for Advertising in Real-Time Bidding Auctions\thanks{\scriptsize This paper was previously circulated under the title ``Online inference for advertising auctions.'' We thank Nan Xu for close collaboration at earlier stages of this paper. We thank Tong Geng, Jun Hao, Xiliang Lin, Lei Wu, Paul Yan, Bo Zhang, Liang Zhang and Lizhou Zheng from \texttt{JD.com} for their collaboration; seminar participants at Cornell Johnson, Berkeley EECS, Stanford GSB: OIT/Marketing, UCSD Rady, Yale SOM and Insper; at the 2019 Conference on Structural Dynamic Models (Chicago Booth), the 2019 Midwest IO Fest, the 2020 Conference on AI/ML and Business Analytics (Temple Fox), the 2020 Marketing Science Conference, the May 2021 QME Rossi Seminar, and the 18th SICS Conference; Mohsen Bayati, Rob Bray, Isa Chaves, Shi Dong, Yoni Gur, Yanjun Han, G\"{u}nter Hitsch, Lalit Jain, Blake McShane, Kanishka Misra, Sanjog Misra, Rob Porter, Adam Smith, Raluca Ursu, Ben Van Roy and Stefan Wager in particular for helpful comments; and Caroline Wang and especially Vitalii Tubdenov for excellent research assistance. Please
contact the authors at \texttt{caio.waisman@kellogg.northwestern.edu} (Waisman), \texttt{hsnair@gmail.com} (Nair) or \texttt{carrion@gatech.edu} (Carrion) for correspondence.}}
\end{spacing}

\author{Caio Waisman \qquad Harikesh S. Nair \qquad Carlos Carrion }

\date{This draft: \today}
\maketitle
\begin{abstract}
\begin{singlespace}
\noindent Real-time bidding systems, which utilize auctions to allocate user impressions to competing advertisers, continue to enjoy success in digital advertising. Assessing the effectiveness of such advertising remains a challenge in research and practice. This paper proposes a new approach to perform causal inference on advertising bought through such mechanisms. Leveraging the economic structure of first- and second-price auctions, we establish novel results that show how the effects of advertising are connected to and hence identified from optimal bids. Importantly, we also outline the precise conditions under which these relationships hold. Since these optimal bids are required to estimate the effects of advertising, we present an adapted Thompson Sampling algorithm to solve a multi-armed bandit problem that succeeds in recovering such bids and, consequently, the effects of advertising, while minimizing the costs of experimentation. We use data from real-time bidding auctions to show that it outperforms commonly used methods to estimate the effects of advertising.
\end{singlespace}

\noindent \vspace{7bp}
\begin{singlespace}
\noindent \textit{Keywords}: Causal inference, multi-armed bandits, advertising auctions
\end{singlespace}
\end{abstract}
\pagebreak{}

\section{Introduction}

\noindent The dominant way of selling ad impressions on ad exchanges (AdXs) is through real-time bidding (RTB) systems, which utilize auctions to allocate user impressions arriving at digital publishers to competing advertisers or intermediaries such as demand-side platforms (DSPs). Most RTB auctions on AdXs are single-unit second-price auctions (SPAs) or single-unit first-price auctions (FPAs). The complexity and scale of available ad inventory, the speed of transactions, and the complex nature of competition imply that advertisers participating in RTB auctions have significant uncertainty about the value of the impressions they are bidding for and the competition they face. Developing accurate and reliable ways of measuring the value of advertising in this environment is essential for advertisers to profitably trade at AdXs and to ensure that acquired ad impressions generate sufficient value. Measurement needs to deliver incremental effects of ads for different types of ad and impression characteristics and needs to be automated. Experimentation thus becomes an attractive way to obtain credible estimates of such causal effects.

Motivated by this, we present a new approach to perform causal inference on RTB advertising in both SPA and FPA settings. Our approach enables learning heterogeneity in the inferred average causal effects across ad and impression characteristics. The novelty of our approach is in addressing the two main challenges that confront developing a scalable experimental design for RTB advertising.

The first challenge is that measuring the average treatment effect ($ATE$) of advertising requires comparing outcomes of users who are exposed to ads with those of users who are not.  A difficulty of the RTB setting is that ad exposure is not under complete control of the experimenter because it is determined by an auction. This precludes the use of simpler experimental designs where ad exposure is directly randomized. Instead, we consider a design in which the experimenter controls only an input to the auction, the bid, but wishes to measure the effect of a stochastic outcome induced by this input, ad exposure. 

The second challenge is managing the cost of experimentation. Obtaining experimental impressions is costly: one has to win an auction to observe the outcome with ad exposure and lose the auction to observe the outcome without it. Costs can be substantial when bidding is not optimized. First, they can emerge from overbidding, in which case the realized cost is high because the paid amount is not compensated by the outcome that is obtained from winning the auction. Second, they can emerge from underbidding, in which case the opportunity cost from losing the auction is high because the outcome that is obtained from losing is lower than the outcome that would have been obtained from paying a high enough bid to win the auction. With potentially millions of auctions in an experiment, suboptimal bidding can make experimentation unviable. Therefore, to be successful an effective experimental design has to deliver inference on the causal effect of ads while also managing the cost of experimentation by implementing a bidding policy that is optimized to the value of the impressions encountered.

It is not obvious how to design an experiment that addresses both challenges simultaneously: optimal bidding requires knowing the value of each impression, whose estimation was the goal of experimentation in the first place. Thus, online methods, which introduce randomization to induce exploration of the value of advertising with concurrent exploitation of the information learned to optimize bids, become very attractive in such settings.

At the core of these online methods is the need to account for the goal of learning the $ATE$ of ad exposure (henceforth called the advertiser's ``inference goal'') and the goal of learning the optimal bidding policy (henceforth called the advertiser's ``economic goal'') concurrently. The tension is that finding the optimal bidding policy does not guarantee proper estimation of ad effects and vice versa. At one extreme, with a bidding policy that delivers on the economic goal, the advertiser could win most of the time, making it difficult to measure ad effects since outcomes with no ad exposures would be scarcely observed. At the other extreme, with pure bid randomization the advertiser could estimate ad effects and deliver on the inference goal but may end up incurring large economic losses in the process.

We contribute by framing the advertiser's problem as a multi-armed bandit (MAB) problem and introducing a statistical learning framework that address both these considerations. In our design, observed heterogeneity is summarized by a context, $x$, bids form arms, and the advertiser's monetary auction payoffs form the rewards, so that the best arm, or optimal bid, maximizes the advertiser’s expected payoff from auction participation given $x$. Exploiting the economic structure of SPAs and FPAs, we outline precise conditions to derive the link between the optimal bid at a given $x$, $b^*(x)$, and the $ATE$ of the ad at the value $x$, or $ATE(x)$. For SPAs, we show that these two objects are equal, so that the twin tasks of learning the optimal bidding policy and estimating ad effects are perfectly aligned. For FPAs, we demonstrate that the two goals are closely related, though only imperfectly aligned. In both cases, we show that the $ATE$s are identified from the optimal bids, so that tackling the economic goal, that is, solving the MAB problem, suffices to estimate ad effects in addition to learning the optimal bidding policy.

To implement our proposed framework, we present a modified Thompson Sampling (TS) algorithm customized to our auction environment trained online via Markov Chain Monte Carlo (MCMC) methods, which we refer to as Bidding Thompson Sampling (BITS). TS is a Bayesian algorithm, which is an attractive way of tackling this problem because it can easily incorporate prior information and flexibly exploit the shared information across arms and contexts by exploiting the full structure of the data. The algorithm adaptively chooses bids across rounds based on current estimates of which bid is the optimal one. These estimates are updated each round via MCMC through Gibbs sampling with data augmentation and the random walk Metropolis-Hastings algorithm. 

Using the iPinYou data set, which contains information on RTB auctions, we show through a series of simulation exercises that our proposed algorithm is able to recover the $ATE$s of advertising and incurs in substantially lower costs of experimentation compared to typical non-adaptive and adaptive approaches. This illustrates the viability of our approach and demonstrates its superior performance against popular competing benchmarks on the economic and inference goals. Importantly, the results indicate that while accounting for correlation in rewards across bids even in a reduced-form way can be helpful in accomplishing the advertiser's goals, further exploiting the structure of the data is more consequential. This suggests that even simpler versions of BITS, which are easier and faster to implement, can be more attractive than commonly used methods to estimate the effects of advertising.

To summarize, the high-level contributions of this paper are twofold. First, it derives explicit mappings between the optimal bids in second-price and first-price auctions and the average treatment effect of ad exposure, thus characterizing the extent of alignment between these objects. It obtains these results by first framing the advertiser's payoff from auction participation as a function of the potential outcomes associated with ad exposure and then by outlining the conditions under which these mappings hold. Importantly, these objects and mappings account for observed heterogeneity across ad impression opportunities. 

Second, it demonstrates how these mappings can be leveraged to run experiments to estimate the expected effect of ad exposure while addressing the costs of experimentation. In particular, it introduces a flexible algorithm, based on Thompson Sampling, that uses MCMC methods to accomplish these tasks concurrently by exploiting the alignment between them.

The rest of the paper discusses the relationship between our approach and the relevant literature and highlights our contributions relative to existing work. The following section defines the experimenting advertiser's problem. Section \ref{sec:balancing-obj} shows how we leverage auction theory to align the advertiser's objectives and Section \ref{sec:achieve} describes the data available to this advertiser. Section \ref{sec:BITS} presents the modified TS algorithm we use to implement our framework. Section \ref{sec:exper} displays results documenting the performance of the proposed algorithm and shows its advantages over competing methods. The last section concludes.

\subsection*{Related literature}

\noindent Our paper lies at the intersection of three fields of study: online learning in ad auctions, experimentation in digital advertising, and causal inference with MABs. Since these streams of literature are mature, we only discuss the studies to which ours is most closely related.

\paragraph{Online learning in ad auctions} \quad This paper relates to the literature on online learning in ad auctions because we solve an online learning-to-bid problem.

Many studies in this literature take the seller's perspective and focus on designing expected profit maximizing mechanisms (e.g., by optimizing reserve prices). However, our study relates more closely to studies that address learning-to-bid problems. For FPAs, \cite{bgmms2019} and \cite{hanZhouWeismannAdverserial2020, hanZhouWeismann2020} present algorithms for contextual bandits that treat the bidder's valuation as the known context, so that there is no need for causal inference since this value corresponds to the treatment effect of the ad. \cite{weed16}, \cite{fj2017}, \cite{FengPodimataSyrgkanis2018}, \cite{mhmsm2020}, and \cite{bgh2021} relax the assumption that advertisers know their valuations before bidding and present algorithms for bid optimization. Except for \cite{mhmsm2020} and \cite{bgh2021}, they also rule out the need for causal inference by assuming that the bidder's valuation is observed when they win the auction, and that the valuation is zero when they lose (as in canonical auction models). This assumption may not be suitable in the context of advertising because one may have a propensity to buy the advertiser's products even in the absence of ad exposure and winning the auction might change this propensity. 

We innovate by allowing the advertiser's valuation to be non-zero even when they lose the auction. We do so by directly inserting the potential outcomes from winning and losing into their payoff function, thereby framing the bid optimization problem in terms of the incremental effect of showing an ad. The idea that this problem should be framed based on incrementality is not novel; see, for example, \cite{xsmlql2016}. However, to our knowledge we are the first to do so formally, both for SPAs and FPAs.\footnote{\cite{xsmlql2016} state without proving a result analogous to our Proposition \ref{prop:alignSPA}, which presents the link between $b^*(x)$ and $ATE(x)$ in SPAs. Since the first version of our paper, \cite{wncx2019}, other papers followed the same approach, such as \cite{mhmsm2020} and \cite{bgh2021}.}

This paper is also related to recent studies that analyze market equilibrium when bidders learn their values by interacting in repeated RTB auctions. Examples include \cite{DikkalaTardos13}, who characterize an equilibrium credit provision strategy for an ad publisher, and \cite{IyerJohariSundararajan14}, who use a mean-field approximation for large markets to study repeated SPAs and apply it to the problem of reserve price optimization. \cite{balseiroBesbesWeintraub2015repeated} characterize equilibrium strategic interactions between budget-constrained advertisers who face no uncertainty about their valuations, but have uncertainty about the number, bids, and budgets of their competitors. \cite{balseiroGur2019learning} and \cite{TunuguntlaHoban21} provide pacing algorithms for repeated SPAs. \cite{TunuguntlaHoban21} also discuss augmenting their algorithm with bandit exploration when the advertiser's valuation has to be learned. Overall, the goals of these papers are different from ours, which is to develop an approach to perform causal inference on the expected effect of ads bought via SPAs and FPAs. 

A relatively smaller literature uses reinforcement learning (RL) approaches to optimize RTB advertising policies (\citealp{cai2017real,wu2018multi, jin2018real}). However, these papers are not concerned with performing causal inference. 

Finally, since we tackle this goal by leveraging key properties of the auction format, we also contribute to a nascent literature on direct applications of auction theory to enable causal inference. While several studies combined experimental designs with auction theory, their goals were to identify optimal policies such as bidding, reserve prices (\citealp{asmt2016,ostrovsky2009reserve,pouget2018optimizing,razk2019}), auction formats (\citealp{chn2016}), or more general mechanisms \citep{kgjs2020}, not to estimate the causal effect of an action such as advertising determined by the outcome of an auction, which is our goal.

\paragraph{Experimentation in digital advertising} \quad This paper relates to the literature on experimental approaches to measure the effect of digital advertising.\footnote{See \cite{GordonetalInefficienciesdigitalAds21} for a detailed review.} One feature that distinguishes our paper from several of these studies is its focus on developing an approach from the advertiser's perspective as opposed to from the publisher's, such as ghost ads for display advertising (\citealp{johnson2017ghost}) or search ads (\citealp{simonovRao2018competition}), which require observation of auction logs or cooperation with the publisher for implementation. In RTB settings, the advertiser bidding on AdXs does not control the auction and does not have access to these logs, precluding the use of such methods. 

Approaches that take the advertiser's perspective include geo-level randomization (\citealp{blakeNoskoTadelis2015consumer}) and randomizing ad campaign frequency caps on DSPs (\citealp{sahni2019experimental}). Unlike these papers, our method leverages bid randomization, is tailored to the RTB setting, is an online inferential procedure, and leverages auction theory for inference. 

\cite{lw2018} also suggest using bid randomization to infer the causal effects of RTB ads. They use bids as an instrumental variable for ad exposure and recover a local average treatment effect, while we leverage auction theory and outline specific conditions to deliver $ATE$s for SPAs and FPAs. In addition, unlike the experimental design we propose, their method is not adaptive and involves pure exploration. Consequently, it does not attempt to minimize the costs of experimentation, which we do. 

Adaptive experimental designs for picking the best creative for ad campaigns are presented in \cite{scott2015multi}, \cite{schwartzetal2017}, \cite{Juetal2019}, and \cite{GengLN20}. The problem addressed in these papers of selecting a best performing variant from a set of candidates is conceptually distinct from the problem addressed in this paper of measuring the causal effect of an RTB ad campaign, and these papers also do not address bidding.

Finally, our approach also relates to that of \cite{fb2018}, who also advocate for experimental designs that are profit maximizing. \cite{fb2018} use a two-period setup to study a treatment decision that is under the experimenter's full control, while we adopt a many-period MAB framework and show how to implement profit maximizing experiments for outcomes over which the advertiser only has imperfect control.

\paragraph{Causal inference with multi-armed bandits} \quad Existing approaches to perform causal inference with MABs differ based on whether they pertain to offline settings, where pre-collected data are available to the analyst, or online settings, where data arrive sequentially, with the latter being relatively more recent. This paper relates more closely to the online stream, so we only discuss papers on online inference.

Performing causal inference with MABs is complicated by the adaptive nature of data collection, wherein future data collection depends on the data already collected. Estimates of arm-specific expected rewards with typical MABs exhibit ``adaptive bias'' \citep{Xu:2013,Villar2015MultiarmedBM}. \cite{Nie2018} showed that archetypal algorithms such as Upper Confidence Bound (UCB) and TS compute estimates of arm-specific expected rewards that are biased downwards. Hence, leveraging MABs for causal inference for RTB ads is hindered even when ad exposure is under the advertiser's complete control. If we set up ad exposure and no ad exposure as arms, so that the difference in rewards between the two arms represents the causal effect of the ad and use a typical algorithm to solve the MAB problem, adaptive bias implies that estimates of the respective arm-specific expected rewards would not yield the $ATE$ of ad exposure. 

Online methods to find the best arm while correcting for adaptive bias include \cite{goldenshluger2013}, \cite{Nie2018}, \cite{BastaniBayati2015}, \cite{dimakopoulou2018estimation}, and \cite{hhzwa2020}.\footnote{Online methods to perform frequentist statistical inference that is valid for MABs, but which avoid correcting for the bias, are presented in \cite{yrjw2017}, \cite{jj2018}, and \cite{Juetal2019}.} These studies aim to estimate without bias the expected reward associated with the best arm. In contrast, our goal is to obtain a consistent estimate of the expected effect of an action (ad exposure) that is imperfectly obtained by pulling arms (placing bids), so that our target treatment whose effect is to be learned is not an arm, but a shared stochastic outcome that arises from pulling arms.\footnote{Hence, arms may be more appropriately thought of as encouragements for treatments, which is analogous to an offline encouragement design (\citealp{ImbensRubin1997}). Thus, our setup also shares similarities with the instrument-armed bandit setup of \cite{Kallus2018IV}, where the treatment-arm pulled may differ from the treatment-arm applied due to non-compliance. However, unlike the setup in \cite{Kallus2018IV}, where exposure to a treatment is the outcome of a choice by the unit to comply, exposure here is determined by a game that is not directly affected by the unit, which characterizes a different exposure mechanism.} In addition, our approach presents a new way of avoiding adaptive bias because the object we seek to estimate and perform inference on relies on the identity of the best arm rather than the expected value of its reward. When a MAB algorithm can correctly recover the identity of the best arm, we can then leverage it for inference on ad effects in an online environment. This is achieved by maintaining a close link to auction theory, which makes our approach different in spirit from the above, more theory-agnostic approaches.

MABs have also been embedded within the structural causal framework of \cite{Pearl2009} in \cite{Bareinboim2etal015}, \cite{Lattimoreetal2016}, and \cite{forneyetal2017counterfactual}. Our paper relates to them as our setting is a specific instance of a structural causal model tailored to the auction setting. As this stream has emphasized, the link to the model in our application is helpful to making progress on the inference problem.\footnote{This approach has also been followed by other papers in economics and marketing that study pricing problems where firms aim to learn the optimal price from a grid of prices, corresponding to arms, which share the same underlying demand function, such as \cite{msa2019}.}

\begin{section}{Problem formulation}

\noindent Our goal is to develop a method to measure the effect of exposing users to ads an advertiser buys programmatically on AdXs. To buy an impression opportunity, advertisers need to participate in an auction ran by the AdX. Winning the auction allows advertisers to display their ads to users. We take the perspective of a focal advertiser that experiments to recover the expected effect of exposing users to their ad.

We define the advertiser's goal of estimating the expected effect of displaying their ad to a user as the \textit{inference goal}. Let $Y(1)$ be the revenue the advertiser receives when their ad is shown to the user and let $Y(0)$ be the revenue they receive when their ad is not shown. Thus, the incremental effect of displaying the ad is $Y(1)-Y(0)$. All the information the advertiser has about the impression opportunity is captured by a variable $x\in\mathbb{X}$.\footnote{In our MAB setup, $x$ is the context of the auction. It can be obtained from a vector of observable display opportunity variables that can include user, impression, and publisher characteristics. For example, if this vector includes the city where the user is located (New York, Los Angeles, or Chicago), the time of day (morning, afternoon, or evening), and the user's age (young, middle-aged, or old), then $x$ can take 27 values, one of which being, for instance, an indicator for Chicago-evening-young.} 

The advertiser's inference goal is to estimate conditional average treatment effects, in which exposure to the ad is the treatment. We denote them by $ATE(x)=\mathbb{E}\left[Y(1)-Y(0)\middle\vert x\right]$. The advertiser needs to estimate this object because they do not have complete knowledge of the distribution of the potential outcomes, $Y(1)$ and $Y(0)$, given $x$. Thus, achieving the inference goal requires the collection of data informative of this distribution.

A method that accomplishes the inference goal has to address four issues, which we discuss below. All four are generated by the distinguishing feature of the AdX environment that the treatment, exposure to the ad, can only be obtained by winning an auction.

\paragraph{Issue 1: The advertiser cannot randomize treatment directly} \quad The first issue is that treatment is determined by an auction over which the advertiser has no complete control; thus, methods that directly randomize treatment are unavailable. However, participation in auctions is under the advertiser's control, so that a viable alternative design randomizes participation across auctions. Random auction participation is analogous to bid randomization, with ``no participation'' corresponding to a bid of zero and ``participation'' corresponding to a positive bid. Nevertheless, as we now show, bid randomization does not suffice to identify causal effects without additional assumptions.

\paragraph{Issue 2: Bid randomization alone is insufficient for identification} \quad The second issue is that bid randomization is not sufficient to identify $ATE$s if the relationship between $Y(1)$, $Y(0)$, and competing bids is unrestricted. To see this, denote the highest bid against which the advertiser competes by $B_{CP}$. For simplicity, assume that $x$ can be ignored. Let $D \equiv \mathbbm{1} \left \{ B_{CP} \leq b \right \}$ denote winning the auction, where $b$ is the advertiser's own bid. The observed outcome, $Y$, is given by $Y = D Y(1) + (1-D) Y(0)$. Let  $Y(1)=\lambda_1+\eta_1$ and $Y(0)=\lambda_0+\eta_0$, where $\lambda_1$ and $\lambda_0$ are constants, and $\eta_1$ and $\eta_0$ are error terms such that $\mathbb{E}[\eta_1]=\mathbb{E}[\eta_0]=0$. Hence, $ATE= \lambda_1 - \lambda_0$. 

Consider estimating the $ATE$ via a simple linear regression. A regression of $Y$ on $D$ using data collected from an experiment in which bids were randomized corresponds to:
\begin{align}\label{eq:regr}
Y_i &=\lambda_0 + ATE \times D_i + \epsilon_i,
\end{align}
where $i$ indexes an observation and $\epsilon_i = D_i \eta_{1i} + (1-D_i) \eta_{0i}$. For the OLS estimator of $ATE$ to be consistent, the indicator $D_i$ has to be uncorrelated with $\eta_{1i}$ and $\eta_{0i}$.\footnote{To be more precise, the actual required condition is the specific case in which $\left [1- \Pr(D=1) \right ] \times \mathbb{E}\left [D \eta_1 \right ]=\Pr(D=1)\times\mathbb{E}\left [D \eta_0 \right ]$. Notice that this condition holds when $D$ is uncorrelated with $\eta_1$ and with $\eta_0$.} There are two potential sources of such correlation, $b$ and $B_{CP}$. Even if $b$ is picked at random, a correlation can still exist through $B_{CP}$. This motivates Assumption \ref{ass:pv} below, which we maintain for the rest of the paper.

\begin{assum}\label{ass:pv} Conditional independence (``private values'')\hfill\break
For all $x\in\mathbb{X}$, $\left \{Y(1),Y(0)  \right \} \independent B_{CP} \vert x $. 
\end{assum}

Assumption \ref{ass:pv} has an auction-theoretic interpretation. It implies that, given $x$, knowledge of $B_{CP}$ has no effect on the advertiser's assessment of $Y(1)$ and $Y(0)$. As we show below, the advertiser's valuation of the impression opportunity equals the treatment effect, so that Assumption \ref{ass:pv} can be seen as analogous to a private values condition.\footnote{We say that it is ``analogous to'' but not exactly a private values condition because the model of ad auctions we present is different from canonical auction models (e.g., \citealp{milgrom1982theory}). In such models, each bidder obtains a signal of the valuation for what they are bidding for prior to the auction and use this signal to form an expectation of this valuation. Under a private values condition, it is without loss of generality to normalize the bidder's valuation to be the signal itself (\citealp{ah2002}, pages 2110--2112). If we applied this formulation here, this would imply that the signal would correspond to the treatment effect itself. This does not fit our empirical setting where the treatment effect is not known to the advertiser prior to bidding, which is the motivation for running the experiment in the first place. Reflecting this, the model we present does not have signals. Consequently, it does not map exactly to the canonical dichotomy between private and interdependent values, which is framed in terms of bidders' signals.} This does not mean that there cannot be correlations with the values that competing advertisers assign to an impression. The assumption is that the common components of bidder valuations are captured by $x$. Part of the motivation arises from the programmatic bidding environment. When advertisers bid in AdXs, they match the ID (cookie/identifier) of the impression with their own private data. If $x$ contains rich information and encompasses what is commonly known about the impression, advertisers' valuations after conditioning on $x$ would only be weakly correlated or not at all. 

An alternative way of interpreting Assumption \ref{ass:pv} is in causal inference terms as an unconfoundedness assumption. Statistically, it is a conditional independence assumption, which is more likely to hold when $x$ captures a lot of information. This is more likely to happen when the experimenter is a large advertiser or an intermediary such as a large DSP, which has access to large amounts of data that can be matched to auctioned impressions in real-time.

Along with bid randomization, Assumption \ref{ass:pv} yields identification of the $ATE$s. From equation (\ref{eq:regr}), OLS can estimate $ATE$ consistently under this condition because both $b$ and $B_{CP}$ become independent of $\eta_{1}$ and $\eta_{0}$. 

\paragraph{Issue 3: High costs of experimentation} \quad  While bid randomization combined with Assumption \ref{ass:pv} is sufficient for identification of $ATE$s, a third issue to consider is the cost of experimentation. This cost can be high under suboptimal bidding. Bidding higher than what is required to win an auction implies a waste of resources due to overbidding, and no bidding involves the opportunity costs from underbidding, especially from users to whom exposure to ads would have been beneficial. In a high frequency RTB environment, a typical experiment can involve millions of auctions, so that if bidding is not properly managed the resulting economic losses can be substantial.

The key to managing these costs is bid optimization, specifically by finding the expected payoff maximizing bid, $b^*(x)$, for all $x$. We refer to the advertiser's goal of learning the optimal bidding policy in the experiment as their \textit{economic goal}. As we discuss below, the inference and economic goals are directly related, though not necessarily perfectly aligned.

To characterize the optimal bid and relate it to the advertiser's inference goal, we first turn to the advertiser's bid optimization problem. We denote the advertiser's payoff as a function of their bid by $\pi \left [ b,Y(1),Y(0),B_{CP} \right ]$. We assume that the advertiser is risk neutral.
The functions $\pi(\cdot)$ for an SPA and an FPA are, respectively:
\begin{align}
	\pi \left [ b,Y(1),Y(0),B_{CP} \right ] &=  \mathbbm{1} \left \{ B_{CP} \leq b \right \} \times \left [ Y(1) - B_{CP} \right ] + \mathbbm{1} \left \{ B_{CP} > b \right \} \times Y(0) \nonumber \\
	&=\mathbbm{1} \left \{ B_{CP} \leq b \right \} \times \left \{ \left [ Y(1) - Y(0) \right ] - B_{CP} \right \} + Y(0) \label{eq:revenueSPA}, \\
 &\text{ and } \nonumber \\
	\pi \left [ b,Y(1),Y(0),B_{CP} \right ] &=  \mathbbm{1} \left \{ B_{CP} \leq b \right \} \times \left [ Y(1) - b \right ] + \mathbbm{1} \left \{ B_{CP} > b \right \} \times Y(0) \nonumber \\
	&=\mathbbm{1} \left \{ B_{CP} \leq b \right \} \times \left \{ \left [ Y(1) - Y(0) \right ] - b \right \} + Y(0).\label{eq:revenueFPA}
\end{align}
Notice that the formulation of auction payoffs in equations (\ref{eq:revenueSPA}) and (\ref{eq:revenueFPA}) is somewhat unconventional. In most auction models, the term $Y(0)$ is set to zero because it is assumed that a bidder only accrues utility when they win the auction. However, this convention is not suitable to our setting given the interpretation of the terms $Y(1)$ and $Y(0)$. A user might have a baseline propensity to purchase the advertiser's product even if they are not exposed to the ad, which is associated with the term $Y(0)$. Exposure to the ad might affect this propensity, implying that $Y(1)\neq Y(0)$. 

The advertiser chooses a bid to maximize their expected payoff from auction participation, $\bar{\pi}(b|x) \equiv \mathbb{E} \left \{ \pi \left [ b,Y(1),Y(0),B_{CP} \right ] \middle \vert x  \right \}$, where the expectation is computed using the joint distribution of $Y(1)$, $Y(0)$, and $B_{CP}$ given $x$. In addition to assuming that the advertiser does not fully know the distribution of $Y(1)$ and $Y(0)$, we assume that they do not know the distribution of $B_{CP}$ as well. This implies that the advertiser faces uncertainty over the joint distribution of $Y(1)$, $Y(0)$, and $B_{CP}$ given $x$, which we denote by $F \left (\cdot, \cdot,\cdot \middle \vert x \right )$. It is important to note that the fact that we postulate that there exists a distribution for $B_{CP}$ does not imply that competitors are randomizing bids or following mixed strategies, although it does allow for it. As in typical game-theoretic approaches to auctions, a given bidder, in this case the advertiser, treats the actions taken by their competitors as random variables, which is why we treat $B_{CP}$ as being drawn from a probability distribution.\footnote{Summarizing competition by $B_{CP}$ is a reduced-form approach. Often $B_{CP}$ can be defined more precisely because more information is available. For example, if the advertiser knew that there were $M$ competitors, then $B_{CP}$ would correspond to the highest order statistic out of the $M$ competing bids. This order statistic could be further characterized by making additional assumptions such as bidder symmetry. If a reserve price was in place, then $B_{CP}$ would further be the maximum between this reserve price and the highest bid. If $M$ were unknown but the advertiser knew the probability distribution governing $M$, then $B_{CP}$ would be this maximum integrated with respect to $M$ using such distribution. The reason why we follow this reduced-form approach is twofold. The first is practical: in settings such as ours, advertisers rarely have information about the number and identities of the competitors they face, so conditioning on or incorporating it would be infeasible. The second is that we are focusing on the optimization problem faced by a single advertiser, who takes the actions of other agents as given. Since $B_{CP}$ can incorporate both a varying number of competitors and a reserve price, it is a convenient modeling device to solve this problem.\label{foot:red}}

Finding $b^*(x)$ when $F \left (\cdot, \cdot,\cdot \middle \vert x \right )$ is unknown is neither standard nor trivial. It is not standard because, under the outlined circumstances, the advertiser faces two levels of uncertainty. The first, lower-order uncertainty is similar to the one faced by bidders in typical auction models: they are uncertain about their competitors' actions, encapsulated by $B_{CP}$. They are also uncertain about their own valuations, because, under this formulation, valuations correspond to treatment effects, which are never observed in practice.

The second, higher-order uncertainty is not present in standard auction models and is the source of the inference goal. While in most auction models bidders do not have complete knowledge of their valuation or of competing bids, they, in equilibrium, know the distributions from which these objects are drawn, which would correspond to the joint distribution of $Y(1)$, $Y(0)$, and $B_{CP}$ given $x$. This is not the case here since the advertiser faces uncertainty regarding $F \left (\cdot, \cdot,\cdot \middle \vert x \right )$. To better see why this bidding problem is not trivial, notice that, without access to data informative of this distribution, bidders would have to integrate over $F \left (\cdot, \cdot,\cdot \middle \vert x \right )$ itself to construct the expected payoff from auction participation. The resulting optimization problem is:
\begin{align}\label{eq:full_opt}
\max_b \mathbb{E}_F \left ( \mathbb{E} \left \{\pi \left [ b,Y(1),Y(0),B_{CP} \right ] \middle \vert F, x \right \} \middle \vert x \right ).
\end{align}
In equation (\ref{eq:full_opt}), the inner expectation is taken with respect to $Y(1)$, $Y(0)$, and $B_{CP}$ conditional on $x$, where $\pi \left [ b,Y(1),Y(0),B_{CP} \right ]$ is given in equations (\ref{eq:revenueSPA}) and (\ref{eq:revenueFPA}). Thus, it reflects the aforementioned lower-order uncertainty. In turn, the outer expectation is taken with respect to $F \left (\cdot, \cdot,\cdot \middle \vert x \right )$ itself, which is reflected in the subscript $F$ and on the conditioning on $F$ in the inner expectation. At the most general level, the advertiser would consider all trivariate probability distribution functions whose support was the three-dimensional positive real line. As such, the optimization problem given in (\ref{eq:full_opt}) is not tractable, and the solution to this problem is in all likelihood highly sensitive to the beliefs over distributions the advertiser can have.

Without access to data, the advertiser would have no choice but to tackle the optimization problem in (\ref{eq:full_opt}). However, as we noted above, achieving the inference goal requires the collection of data, which can also be used to address the economic goal. Therefore, instead of tackling the optimization problem in (\ref{eq:full_opt}), one strategy to optimize bidding would be to use the data collected in the experiment to construct updated estimates of $\mathbb{E} \left \{ \pi \left [ b, Y(1),Y(0),B_{CP} \right ] \middle \vert F, x \right \}$ and optimize bids with respect to this estimate as the experiment progressed. This way, the data generated from the experiment would be used to address the advertiser's inference goal and to optimize expected profits to address their economic goal, and thus both goals would be addressed concurrently. 

\paragraph{Issue 4: Aligning the inference and economic goals}\quad The final issue is aligning the simultaneous pursuit of the inference and economic goals. This issue arises because typical strategies aimed at tackling one of the goals can possibly have negative impacts on accomplishing the other, suggesting a tension between the two. 

To see this, consider what would happen if the experiment focused solely on the advertiser's inference goal by randomizing bids without any concurrent bid optimization. We already alluded to the consequences of this for the advertiser's economic goal: pure bid randomization can hurt the economic goal by inducing costs from suboptimal bidding.

Consider now what would happen if the experiment focused only on the advertiser's economic goal. The payoffs from bidding $b$ are stochastic from the advertiser's perspective, and the optimal bid is the maximizer of the expected payoff from auction participation, $\bar{\pi}(b|x)$, which is unknown to the advertiser. Hence, pursuing the economic goal involves finding the best bid in an environment where payoffs are stochastic and maximizing expected payoffs against a distribution which has to be learned by exploration. MAB and RL approaches are thus attractive in this situation because they can recover $b^*(x)$ while minimizing the costs from suboptimal bidding, which pure randomization does not account for. By following one such strategy, the advertiser would adaptively collect data to learn a good bidding policy by continuously re-estimating and re-optimizing $\bar{\pi}(b|x)$. 

In principle, these data could also be used to estimate $ATE(x)$ by running the regression in (\ref{eq:regr}) for each $x$, for example. However, the adaptive nature of the data collection procedure induces autocorrelation in the data, which can impact statistical properties of estimators. Moreover, even if all desired properties hold, underlying characteristics of the data and the algorithm used can affect the estimator adversely. To see this, consider the following example. For simplicity, assume once again that $x$ can be ignored and further assume that $\Pr(B_{CP} \leq b^*)\approx 1$. A successful algorithm could converge quickly, eventually yielding relatively few observations of $Y(0)$ compared to of $Y(1)$, which would hinder the inference goal since typical estimators of $ATE$s based on such imbalanced data tend to be noisy.

This illustrates that the inference and economic goals can possibly be at odds despite being clearly related since they both depend on knowledge of the distribution $F \left (\cdot,\cdot,\cdot\middle\vert\cdot \right )$. The challenge faced by the advertiser to accomplish these goals is that this distribution is unknown, and while there are several approaches to gather data and tackle each goal in isolation, it is unclear whether they can perform satisfactorily in achieving both goals concurrently. We present a method to solve this challenge next.

\end{section}

\begin{section}{Aligning the advertiser's objectives}\label{sec:balancing-obj}

\noindent We align the inference and economic goals by leveraging the structure of the advertiser's bid optimization problem, which shows how these goals are linked. Because bidding depends on the auction format, the extent to which the two goals are aligned differs across auction formats. We show that in SPAs the inference and economic goals can be perfectly aligned, while in FPAs leveraging this linkage is still helpful, but the goals can only be imperfectly aligned. 

To characterize our approach, we consider the limiting outcome of maximizing the true expected profit function with respect to bids when the joint distribution $F \left (\cdot, \cdot,\cdot \middle \vert \cdot \right )$ is known. In what follows, we use the expressions in equations (\ref{eq:revenueSPA}) and (\ref{eq:revenueFPA}) ignoring the second term, $Y(0)$, since it does not depend on the advertiser's bid. This expression also has the benefit of directly connecting the potential outcomes to this auction-theoretic setting, with the treatment effect, $Y(1) - Y(0)$, taking the role of the advertiser's valuation.

We can write the objects the advertiser aims to learn in an SPA and an FPA as the maximizers with respect to $b$ of, respectively,
\begin{align}
	 \bar{\pi}(b \vert x) &= \Pr \left ( B_{CP} \leq b \middle \vert x \right ) \times \mathbb{E} \left \{ \left [ Y(1) - Y(0) \right ] - B_{CP} \middle \vert B_{CP} \leq b; x \right \}, \label{eq:auc_epaySPA} \\ 
  &\text{ and } \nonumber \\
  	 \bar{\pi}(b \vert x) &= \Pr \left ( B_{CP} \leq b \middle \vert x \right ) \times \mathbb{E} \left \{ \left [ Y(1) - Y(0) \right ] - b \middle \vert B_{CP} \leq b; x \right \}. \label{eq:auc_epayFPA}
\end{align}
The expectations in equations (\ref{eq:auc_epaySPA}) and (\ref{eq:auc_epayFPA}) are taken with respect to $Y(1)$, $Y(0)$, and $B_{CP}$. The conditioning on the distribution $F$ is omitted since $\bar{\pi} \left ( \cdot \middle \vert x \right )$ is the true expected profit function, so it uses the true $F \left ( \cdot, \cdot, \cdot \middle \vert \cdot \right )$ to compute the relevant expectations and probabilities. We denote the maximizers of these respective expressions by $b^*(x)$. We ensure that $b^*(x)$ is well-defined and unique with the following assumption.
\begin{assum}\label{ass:joint} {Properties of $F\left ( \cdot, \cdot, \cdot \middle \vert \cdot \right )$} \hfill\break
For all $x\in\mathbb{X}$: \\
\subasu The joint distribution $F(\cdot,\cdot,\cdot \vert x)$ admits a density, $f(\cdot,\cdot,\cdot\vert x)$. \label{ass:dens} \\
\subasu $\mathbb{E} \left [Y(1) \middle \vert x \right ]<\infty$, $\mathbb{E} \left [Y(0) \middle \vert x \right ]<\infty$, and $\mathbb{E} \left [B_{CP} \middle \vert x \right ]<\infty$. \label{ass:fin_means} \\
\subasu If $ATE(x)>0$ then there exists an interval $ \left (\underline{ATE}(x), \overline{ATE}(x) \right )$ such that $0 < \underline{ATE}(x) < ATE(x) < \overline{ATE}(x) < \infty$ in which the density of $B_{CP}$ given $x$, $f_{CP}\left ( \cdot \middle \vert x \right )$, is strictly positive. \label{ass:pos_dens} \\
\subasu If $ATE(x)>0$ the conditional reversed hazard rate of $B_{CP}$ given $x$, $\frac{f_{CP}\left(b_{CP}|x\right)}{F_{CP}\left(b_{CP}|x\right)}$, is decreasing in $b_{CP}$ in $ \left (\underline{ATE}(x), \overline{ATE}(x) \right )$ as defined above. \label{ass:rev_haz}
\end{assum}

Assumption \ref{ass:joint} gives conditions on $F \left ( \cdot, \cdot, \cdot \middle \vert \cdot \right )$ required for us to establish the results presented below. These conditions are mild and relatively common in auction models. Assumption \ref{ass:dens} is standard and made for tractability. In turn, Assumption \ref{ass:fin_means} ensures that the expressions given in equations (\ref{eq:auc_epaySPA}) and (\ref{eq:auc_epayFPA}) are well-defined. 

Assumption \ref{ass:pos_dens} is required for us to establish that the optimal bids are unique. Notice that this condition is equivalent to an overlap assumption that $P(D = 1|x) \in (0,1)$, where $P(D = 1|x)$ is the propensity score. Because $D \equiv \mathbbm{1} \left \{ B_{CP} \leq b \right \}$ and $P(D = 1|x) \equiv P(B_{CP} \leq b|x)$, Assumption \ref{ass:pos_dens} implies that $P(B_{CP} \leq b|x) \in (0,1)$.

Finally, Assumption \ref{ass:rev_haz} is only required to determine $b^*(x)$ for FPAs and is a sufficient condition for $b^*(x)$ to be unique. It states that the distribution of $B_{CP}$ conditional on $x$ has a decreasing reversed hazard rate in $ \left (\underline{ATE}(x), \overline{ATE}(x) \right )$. This property holds for several distributions, including all decreasing hazard rate distributions and increasing hazard rate Weibull, gamma, and lognormal distributions.\footnote{For a comprehensive discussion about reversed hazard rate functions, see, for example, \cite{block_savits_singh_1998}.}

We now investigate the relationship between $b^*(\cdot)$ and $ATE(\cdot)$ under Assumptions \ref{ass:pv} and \ref{ass:joint}. We present this relationship first for SPAs and then for FPAs, followed by a discussion about their novelty and implications. To our knowledge, ours was the first paper that obtained these links between optimal bids and $ATE$s.\footnote{\cite{xsmlql2016} stated the result of Proposition \ref{prop:alignSPA} without proving it. In turn, \cite{mhmsm2020} provided a more general version that encompasses both Propositions \ref{prop:alignSPA} and \ref{prop:alignFPA} without formally proving it. After the first version of our paper, \cite{wncx2019}, \cite{bgh2021} demonstrated analogous results to ours.} 

\begin{prop}Optimal bid in SPAs\label{prop:alignSPA}\hfill\break
\normalfont Suppose that Assumptions \ref{ass:pv}, \ref{ass:dens}, \ref{ass:fin_means}, and \ref{ass:pos_dens} hold. If the auction is an SPA, then $b^*(x)= \max \left \{ 0, ATE(x) \right \}$.
\end{prop}

Define $\chi\left(b\middle \vert x\right)=b+\frac{F_{CP}\left(b|x\right)}{f_{CP}\left(b|x\right)}$.

\begin{prop}Optimal bid in FPAs\label{prop:alignFPA}\hfill\break
\normalfont Suppose that Assumptions \ref{ass:pv} and \ref{ass:joint} hold. If the auction is an FPA, it then follows that $b^*(x)= \max \left\{0,\chi^{-1}\left [ATE(x) \middle \vert x \right ] \right \}$. 
\end{prop}

\paragraph{Novelty and implications} \quad Our main modeling innovation is to insert the potential outcomes, $Y(1)$ and $Y(0)$, directly into the bidder's payoff function.\footnote{Since the first version of our paper, \cite{wncx2019}, this approach was then followed by \cite{mhmsm2020}.} This leads us to the key result that the bidder's valuation equals the treatment effect, $Y(1)-Y(0)$. This has crucial consequences for characterizing the optimal bids in terms of $ATE$s, for determining which algorithms can be used to solve the MAB problem to recover the optimal bids, and for obtaining estimators of the $ATEs$. We describe them below.

The proofs, which are given in Appendix \ref{app:proofs}, rely on first representing the advertiser's assessment of their valuations as $ATE(x)$, which is possible due to Assumptions \ref{ass:pv}, \ref{ass:dens}, and \ref{ass:fin_means}. Assumption \ref{ass:pos_dens} allows us to find the optimal bid for SPAs by inspecting the objective function; in turn, Assumptions \ref{ass:pos_dens} and \ref{ass:rev_haz} allow us to determine the optimal bid for FPAs and its uniqueness through the first-order condition.

The relationships derived here are new and not off-the-shelf results from auction theory. Since potential outcomes had not been previously embedded into auction models, the result that valuations equal treatment effects and its relation to $b^*(x)$ had to be derived from first principles. For instance, while assuming that valuations are positive is innocuous in standard auction models, assuming it in our setting is not. Setting $Y(1)\geq Y(0)$ corresponds to assuming monotone treatment response \citep{manski1997}. From a causal inference perspective, this is not a trivial assumption, which is why we do not maintain it. When characterizing $b^*(x)$, we thus have to account for the possibility that valuations are negative, leading to the results that the optimal bids are equal to the maximum between zero and what indeed are objects analogous to those from standard auction theory.

Another important feature is the role of Assumptions \ref{ass:pos_dens} and \ref{ass:rev_haz}, which guarantee uniqueness of $b^*(x)$. This is especially relevant in the context of SPAs. The standard result is that bidding one's valuation is a \textit{weakly} dominant strategy, therefore allowing for the possibility of infinitely many optimal bids, which can be especially problematic in the context of solving a MAB problem. This is an often overlooked issue and is why we make Assumption \ref{ass:pos_dens}. We believe that explicitly outlining these conditions that guarantee the uniqueness of optimal bids is a contribution, even if minor.

Our insight that valuations equal treatment effects has another important consequence: due to the ``fundamental problem of causal inference'' \citep{holland1986}, valuations are never perfectly observed. Consequently, most existing learning-to-bid algorithms cannot be applied in our setting, because, as noted in the literature review, they assume that valuations are observed, obviating the need for causal inference. 

Finally, the equivalence between valuations and treatment effects and the consequent relationships between $b^*(x)$ and $ATE(x)$ act as identification results, which we leverage to recover the $ATE$s. In particular, they show how we can recover $ATE(x)$ by simply recovering $b^*(x)$ in SPAs, and by recovering $b^*(x)$ and the reverse hazard rate of the distribution of $B_{CP}$ evaluated at $b^*(x)$ for FPAs. Importantly, these results imply that we can recover the $ATE$s from the identity of the best arm instead of from the expected reward associated with the optimal arm. The latter is the usual estimator employed in the literature that merges causal inference with MAB problems. Such estimators suffer from a specific type of adaptive bias (see, for example, \citealp{Nie2018}), which we are therefore able to avoid.

\end{section}

\begin{section}{Accomplishing the advertiser's objectives}\label{sec:achieve}

\noindent We leverage Propositions \ref{prop:alignSPA} and \ref{prop:alignFPA} to develop a method that concurrently accomplishes the advertiser's goals. Our proposal is an adaptive approach that learns $b^*(\cdot)$ over a sequence of display opportunities. 

\subsection{Data generating process}\label{ssec:dgp}

We begin by stating the following assumption, which we maintain throughout.

\begin{assum}\label{ass:dgp}Independent and identically distributed (i.i.d.) data\hfill\break
$\left \{Y_i(1),Y_i(0),B_{CP,i}  \right \} \overset{\text{i.i.d.}}{\sim} F(\cdot,\cdot,\cdot \vert x_i)$ and $x_i \overset{\text{i.i.d.}}{\sim} F_x(\cdot)$.
\end{assum}

Assumption \ref{ass:dgp} is typically maintained in stochastic bandit problems: that the randomness in payoffs is $i.i.d.$ across occurrences of play. It imposes restrictions on the data generating process (DGP). For instance, if competing bidders solved a dynamic problem because of longer-term dependencies, or budget or impression constraints, $B_{CP}$ could become serially correlated, in which case this condition would not hold. 

A different concern regards the context, $x$. Given recent developments in privacy policies and regulation, we treat each observation $i$ as an impression opportunity because these policies can limit and even prevent the use of cookies, which enable advertisers to track the same user over time. This can be problematic since users' responses can be altered by the number of times they are exposed to the ad. 

If the same user can be tracked, a simple way of addressing this concern would be to add the number of previous impressions to $x$, but doing so would violate Assumption \ref{ass:dgp} because $x$ would then become serially correlated. As we show in Section \ref{sec:BITS}, our algorithm would ignore this autocorrelation, which would yield a departure from more standard MAB models. A more rigorous approach would incorporate this correlation directly into the algorithm, which is the route \cite{bgh2021} followed. We view combining their approach with the algorithm we outline as an interesting avenue for future research.

It is important to note that Assumption \ref{ass:dgp} is not required for Propositions \ref{prop:alignSPA} and \ref{prop:alignFPA} to hold, which guarantee a well-defined and unique optimal bid across impression opportunities. We maintain this assumption for convenience because it enables us to cast our problem as a nonlinear stochastic bandit problem.
Furthermore, given our algorithm of choice, we perform Bayesian inference based on the posterior distribution obtained from the data, which, as we argue below, incorporates the correlation in these data, including that of $x$. Therefore, the unit of observation, $i$, can be viewed as a user, and the context, $x$, can contain previous displays, frequencies of displays, as well as other associated variables.

\subsection{Observed data}\label{ssec:ob_data}

Algorithms used to solve MAB problems would base the decision of which bid to place in round $t$, $b_t$, on a tradeoff between randomly picking a bid to obtain more information about its associated payoff (exploration) and the information gathered until then on the optimality of each bid (exploitation). The information at the beginning of round $t$ is a function of all data collected until then, which we denote by $H_{t-1}$. Each observation $i$ in these data is an ad auction but can also be a user that can be tracked.

For the analysis of SPAs, it is useful to define the variable $\bar{B}_{CP}\equiv\min\{B_{CP},b\}$. We can write $H_{t}=\left \{ b_{\tau},x_{\tau}, D_{\tau}, Y_{\tau}, \bar{B}_{CP,\tau}, \omega_{\tau} \right \}_{\tau=1}^{t}$ for SPAs and $H_{t}=\left \{ b_{\tau},x_{\tau}, D_{\tau}, Y_{\tau}, \omega_{\tau} \right \}_{\tau=1}^{t}$ for FPAs. In both cases, the $\omega$s are seeds, independent from all other variables, required for randomization depending on which algorithm is used.

These data suffer from two issues. The first, common to both auction formats, is the ``fundamental problem of causal inference'' \citep{holland1986}: $Y(1)$ and $Y(0)$ are never observed at the same time. The second concerns what we observe regarding $B_{CP}$ and differs across the two auction formats. For SPAs, we have a censoring problem that arises from the competitive environment: $B_{CP}$ is only observed when the advertiser wins the auction; otherwise, all they know is that it was larger than $b$. Hence, the observed data have a similar structure to the one from the Type 4 Tobit model as defined by \cite{amemyia}. However, for FPAs this restriction is stronger: we never observe $B_{CP}$ and only have either a lower or upper bound on it depending on whether the advertiser wins the auction, so that the observed data have a Type 5 Tobit model structure.

\end{section}

\begin{section}{Bidding Thompson Sampling (BITS) algorithm}\label{sec:BITS}

\noindent We now propose a specific procedure to achieve the advertiser's goals, which is a version of the TS algorithm. We refer to it as Bidding Thompson Sampling (BITS). We proceed in four steps. First, we given a general description of how the algorithm operates. Second, we outline and discuss the specific parametrization we adopt. Third, we describe how we compute estimates of $b^*(x)$ and $ATE(x)$. Fourth, we address how one can perform inference on the $ATE$s. Finally, we briefly discuss the general main challenges to implementing this sort of algorithm in practice. We further discuss several generalizations that can be made and relate them to our current approach in Appendix \ref{app:add}.

\begin{subsection}{General procedure}

\noindent It is not our goal to solve for or implement the optimal learning policy that minimizes regret over a finite number of rounds of play. To our knowledge, a general optimal solution for contextual MAB problems with arbitrary nonlinear expected rewards as a function of contexts, arms, and parameters, and correlated rewards across contexts and arms, such as the one we consider, is not yet known. Instead, we require an algorithm that can easily accommodate and account for information shared across actions. Hence, we make use of the TS algorithm \citep{thompson1933}, which is a Bayesian heuristic to solve MAB problems.\footnote{See \cite{scott2015multi} for an application to computational advertising and \cite{russo2018} for a survey.} 

The TS algorithm often starts by parametrizing the distributions of rewards associated with each arm. Since in our problem we treat $b$ as continuous and the DGP behind all actions, the distribution $F(\cdot,\cdot,\cdot\vert\cdot)$, is the same, we choose to parametrize it instead. Denote our vector of parameters of interest by $\theta$. Expected rewards depend on $\theta$, so we will often write $\bar{\pi}(\cdot \vert \cdot,\theta)$. The algorithm runs while a criterion, $c_t$, is below a threshold, $T$. At the end of round $t$, the prior over $\theta$ is updated by the likelihood of all the data, $H_t$. We denote the number of observations gathered on round $t$ by $n_{t}$ and the total number of observations gathered by the end of round $t$ by $N_t=\sum_{\tau=1}^t n_{\tau}$. If $n_{t}=1$ for all $t$, the algorithm proceeds auction by auction. We present it in this way to accommodate batch updates. Given the posterior distribution of $\theta$ given $H_t$, we compute the posterior expected payoff function, $\bar{\pi}_t (\cdot|\cdot,\theta)$ and update the criterion $c_t$. In round $t+1$, we place the bid $b_t(\cdot)\equiv \arg\max_{b\in\mathbb{R}_+}\bar{\pi}_t (\cdot|\cdot,\theta)$.

\end{subsection}

\begin{subsection}{Specific parametrization}\label{sec:par_DGP}

\noindent Given the structure of the data, our algorithm requires re-implementing a Bayesian estimator of a Type 4 or Type 5 Tobit model on each round. The parametric structure we impose is necessary for the implementation of the algorithm but does not represent formal assumptions we make about the true DGP. Thus, we choose a structure that is flexible enough for us to remain agnostic about the true DGP but that is also sufficiently tractable. 

Let $X_i$ be a $P$-dimensional vector interchangeable with $x_i$. We assume that:
\begin{align}\label{eq:normal}
\begin{split}
\begin{bmatrix}\log Y_i(1) \\ \log Y_i(0) \end{bmatrix} \Big\vert X_i &\overset{i.i.d.}{\sim} \sum_{k=1}^{K} \xi_{k} \times N \left ( \begin{bmatrix} X_i'\delta_{k,1} \\ X_i'\delta_{k,0} \end{bmatrix}, \begin{bmatrix} \sigma_{k,1}^2 & \rho_k \sigma_{1,k} \sigma_{0,k} \\ \rho_k \sigma_{1,k} \sigma_{0,k} & \sigma_{0,k}^2 \end{bmatrix}  \right )\\&\equiv \sum_{k=1}^{K} \xi_{k} \times N \left (\Delta_k'X_i, \Sigma_k \right ) \\
 \log B_{CP,i} \Big\vert X_i &\overset{i.i.d.}{\sim} \text{Gumbel} \left (X_i'\delta_{CP}, \sigma_{CP} \right )
\end{split}
\end{align}
where $\Delta_k\equiv \left [\delta_{1,k}, \delta_{0,k} \right ]$. Hence, we use a finite mixture of lognormal distributions for $ \left [Y(1),Y(0) \right ]'$ with $K$ components and mixture weights $\xi_{k}$ and an independent Gumbel distribution with location $X_i'\delta_{CP}$ and scale $\sigma_{CP}$ for $\log B_{CP}$. We define $\theta_{k,Y} \equiv \left [\text{vec}(\Delta_k),\sigma_{k,1}^2, \sigma_{k,0}^2,\rho_k, \xi_{k} \right ]'$, $\theta_{CP} \equiv \left [\delta_{CP}',\sigma_{kP}\right ]'$, and $\theta \equiv \left [ \left \{\theta_{k,Y} \right \}_{k=1}^{K}, \theta_{CP} \right ]'$. Notice that this parametrization imposes Assumptions \ref{ass:pv} and \ref{ass:joint}.

These choices of distributions warrant additional comments. We use a mixture of lognormal distributions for $ \left [Y(1),Y(0) \right ]'$ solely for convenience. Because these variables are measured in monetary amounts it is attractive to model them in a flexible manner, which this mixture approach allows us to do. In turn, the Gumbel distribution can be motivated by auction and extreme value theories. Consider a setting with $M$ $i.i.d.$ random variables. Extreme value theory tells us that the distribution of the maximum of these variables converges to a Fr\'{e}chet under appropriate mild conditions as $M\to\infty$. Thus, if we are willing to assume that the advertiser faces many competitors, which is often true in RTB settings \citep{balseiroBesbesWeintraub2015repeated,balseiroGur2019learning}, and that the bids submitted by these advertisers are $i.i.d.$, which could be justified by an independent private values (IPV) assumption, then the Fr\'{e}chet could provide a good approximation to the distribution of $B_{CP}$, so that a Gumbel can approximate that of $\log B_{CP}$ well. We emphasize that we do not explicitly maintain such assumptions, but rather rely on such conditions to provide some theoretical backing for this parametrization. 

\end{subsection} 

\begin{subsection}{Computing $\pmb{ATE}$s and optimal bids}\label{ssec:comput}

\subsubsection{Drawing from posterior distributions}\label{sssec:post_summ}

Our algorithm requires us to obtain draws from the posterior distribution of $\theta$ given $H_t$ at the end of each round $t$. We now provide a brief description of the MCMC procedures we use to obtain such draws. These procedures are presented in detail in Appendix \ref{App:full-cond-gibbs}.

Obtaining draws of $\theta_Y \equiv \left [\theta_{k,Y} \right ]_{k=1}^{K}$ is straightforward but for two details. First, the parameters $\rho_k$ are not point identified without further assumptions because $Y(1)$ and $Y(0)$ are never observed at the same time. While there exist methods that attempt to obtain information about these correlation coefficients in such situations,\footnote{For example, \cite{vijve1993} exploited the semi-definiteness of $\Sigma$ to obtain bounds on $\rho$.} we choose to obtain draws of $\theta_{Y,1} \equiv \left [\delta_{k,1}, \sigma_{k,1}^2, \xi_{k,1} \right ]_{k=1}^{K}$ separately from draws of $\theta_{Y,0} \equiv \left [\delta_{k,0}, \sigma_{k,0}^2, \xi_{k,0} \right ]_{k=1}^{K}$, which can be interpreted as treating the potential outcomes as independent from each other. 

The reason why we follow this approach is twofold. First, the correlation coefficients are not crucial to our analysis as they do not affect the $ATE$s. In addition, as equations (\ref{eq:revenueSPA}) and (\ref{eq:revenueFPA}) illustrate, the distributions of rewards do not depend on these coefficients. Second, this approach has precedent in the literature \citep{ch2000}. Nevertheless, in our simulation exercises we consider a DGP in which $Y(1)$ and $Y(0)$ are not independent. As we show in Section \ref{sec:exper}, ignoring this correlation does not impact our method adversely.

The second challenge in obtaining draws of $\theta_Y \equiv \left [\theta_{k,Y} \right ]_{k=1}^{K}$ are the missing data. Nevertheless, Assumption \ref{ass:pv} effectively implies missingness at random, so that drawing the missing values and augmenting the data becomes straightforward. Therefore, the procedure to obtain draws of $\theta_{Y,1}$ and $\theta_{Y,0}$ from their posterior distributions given the data consists of the typical Gibbs sampling method used to estimate finite mixtures of normal distributions with an additional simple data augmentation step, akin to \cite{ac1993}. 

The procedure to obtain draws of $\theta_{CP}$ is arguably more complex. Because we cannot obtain conditional conjugacy under the Fr\'{e}chet distribution, and thus use Gibbs sampling, we obtain draws from the posterior distribution of $\theta_{CP}$ given the data using a random walk Metropolis-Hastings procedure instead. We implement its standard version using a multivariate normal proposal distribution. 

\subsubsection{Estimating $\pmb{ATE}$s}

Given $Q$ draws from the posterior distribution of $\theta$ given $H_t$, using (\ref{eq:normal}) the estimator of $ATE(X)$ at the end of round $t$ is:
\begin{align}\label{eq:ate_est}
    \widehat{ATE}_t(X)=\frac{1}{Q}\sum_{q=1}^Q  \sum_{k=1}^{K} \left [ \xi_{k,1}^{(q)}\exp \left \{ X'\delta_{k,1}^{(q)} + 0.5 \sigma_{k,1}^{2,(q)} \right \} - \xi_{k,0}^{(q)}\exp \left \{ X'\delta_{k,0}^{(q)} + 0.5 \sigma_{k,0}^{2,(q)} \right \} \right ].
\end{align}

\subsubsection{Estimating optimal bids}

Using Proposition \ref{prop:alignSPA}, it follows that $b_t(X)=\max \{0, \widehat{ATE}_t(X) \}$ for SPAs. As a result, implementing our algorithm in the context of SPAs does not require us to estimate and update $\theta_{CP}$. This not only greatly simplifies the method but also highlights its flexibility: bids are treated as continuous, and the only parametric assumption made corresponds to the finite mixture of normal distributions, which is a flexible model that treats the distribution of potential outcomes in a somewhat agnostic fashion. 

For FPAs, using Proposition \ref{prop:alignFPA} and (\ref{eq:normal}) we can write the estimated bidding function at the end of round $t$ as:
\begin{align}\label{eq:est_bid_fun}
    \hat{\chi}_t(b|X)=b+\frac{1}{Q}\sum_{q=1}^Q\sigma_{CP}^{(q)}b^{1+\frac{1}{\sigma_{CP}^{(q)}}} \exp \left \{-\frac{X'\delta_{CP}^{(q)}}{\sigma_{CP}^{(q)}} \right \}, 
\end{align}
so that the optimal bid at the end of round $t$ is $b_t(x)= \max \left \{ 0, \hat{\chi}_t^{-1} \left [ \widehat{ATE}_t(X) |X \right ] \right \}$, where $\widehat{ATE}_t(X)$ is given in equation (\ref{eq:ate_est}). Notice that $\hat{\chi}_t(\cdot|X)$ is strictly increasing and so its inverse is well-defined. It is also important to note that, under FPAs, in practice we need to normalize $\sigma_{CP}$ since we never observe $B_{CP}$.

\end{subsection}

\subsection{Bayesian inference}

\noindent Given that our algorithm leverages a Bayesian estimator to solve the MAB problem, it is convenient to use Bayesian inference to assess the uncertainty around the estimates of the $ATE$s. Under this framework, the uncertainty around these estimates is captured through the posterior probabilities associated with them. This approach does not resort to asymptotic properties or approximations, although standard results can be invoked to establish the typical asymptotic properties of the resulting Bayes estimators. 

One advantage of adopting Bayesian inference is that it accommodates the serial correlation in the data in a straightforward way through the likelihood function. Let the likelihood function of the data be $\ell \left (H_1, H_2,...,H_t,...,H_T \middle \vert \theta \right )$. We can factorize it as 
\begin{align}\label{eq:likel_factor}
\ell \left (H_1,...,H_T \middle \vert \theta \right )&=\ell \left (H_T \middle \vert H_1,...,H_{T-1}; \theta \right ) \times \ell \left (H_{T-1} \middle \vert H_1,...,H_{T-2}; \theta \right ) \times ... \times \ell \left (H_1 \middle \vert  \theta \right ).
\end{align}
Given how our algorithm works, conditional on all data collected until round $t-1$, the data from round $t$, $H_t$, are distributed based on the resulting likelihood function given in (\ref{eq:normal}). This is because the data collected until round $t-1$ directly determines the advertiser's own bids and are independent from the remaining random variables drawn in round $t$. Consequently, the autocorrelation implied by the adaptive collection of the data is captured in the posterior distribution.

\subsection{Challenges for practical implementation}\label{sec:prac_imp}

There are three big challenges with implementing this sort of algorithm in practice:
\begin{enumerate}

    \item Observing $Y(0)$: In most cases, ad platforms only log realized outcomes conditional on exposure (e.g., clicks). However, as we noted above, there has been little consideration about collecting data about $Y(0)$ or the potential outcome when a bid was not submitted but the user still pursued a purchase or perhaps even no purchase.  

    \item Delayed feedback: We are also interested in settings the tech industry refers to as ``deep conversions'' (i.e., purchases) as opposed to what it refers to as ``shallow conversions'' (i.e., page views). However, purchases generally occur less often than page views and thus this feedback takes time to realize, and it is critical for a dynamic algorithm to have an``estimate'' of these decisions in order to proceed with the dynamic allocation and optimization. It can be important to define an attribution model for how exposures, clicks, and conversions relate to one another.
    
    \item We model potential outcomes as representing purchase events or events whose outcomes can be expressed in monetary amounts in order to be compatible with the monetary nature of bidding. In the display ads industry, these may be known as the value of a particular action, so other actions besides purchases may be accommodated as long as they can me measured in monetary amounts.
    
\end{enumerate}

The first challenge may be overcome by defining the appropriate metric to represent the outcome and also by having in place the infrastructure to log the outcomes, contexts, submitted bids, and competitor bid information needed for the algorithm. We expect the algorithm to be used by an advertiser who can define purchase or even conversions rates with respect to its brand as a whole or a subset of its products in a period of time for the campaign of interest; this period of time is usually during the execution of the algorithm (e.g., a week or multiple weeks). This metric should be well defined for $Y(1)$ and $Y(0)$.

The second challenge encompasses at least two sub-challenges concurrently: applying an attribution model for exposure to conversion and estimating the conversion rate given the presence of delay. The first issue refers to how to deal with the problem of multiple exposures, multiple clicks, and so on as they relate to conversions. As \cite{chapelle2014} explains, one strategy is to use exposures and clicks prior to conversion in a time window; in particular, they mention 30 days as a possibility, where conversions beyond 30 days are ignored, and also the use of a last attribution model to assign the conversion to the last exposure, last click, first conversion, while the rest are disregarded. This may also be applied for the lack of exposure to last click and first conversion within the time window through a timestamp matching exercise. For the issue of delay, the delay distribution along with the conversion rate can be jointly modeled as \cite{chapelle2014} discusses. In addition, the weighting scheme proposed in \cite{wang2022} for implementation in a MAB algorithm can be similarly adapted for the proposed algorithm. The key intuition in \cite{wang2022} is that the delay distribution may be used to weight the observations to correct for the fact that each observation have different likelihood of converting depending on the delay elapsed so far. Moreover, it should be noted that most TS-MAB algorithms implemented in display advertising do not update the posterior distributions in real-time, and rather use batches where serving is always real-time but reward estimation is done every fixed amount of time (e.g., from 20 to 40 minutes), thus granting time for data preparation. Hence, combining the attribution model with the modeling of delay and conversion for reward estimation can proceed with the selected data. 

Finally, the third challenge requires converting the conversion rates to a monetary metric, which may be achieved by multiplying the conversion rates with the average order value per conversion (or even a look-up table of these values) for the products of the advertiser or the relevant subset in the advertising campaign. More sophisticated approaches may be pursued such as using auxiliary data to estimate the value of a purchase conditional on a conversion, and then embedding these estimates in the profit function along with the conversion rate.

\end{section}

\begin{section}{Empirical evaluation}\label{sec:exper}

\noindent In this section, we evaluate the performance of BITS through simulations based on a DGP calibrated using real RTB data. We first describe how this DGP is calibrated, followed by a description of the methods to which we compare BITS and of the criteria we use to make the comparisons. Finally, we present the results of these exercises.

\subsection{The iPinYou data set and DGP calibration}

\noindent We use the iPinYou data set for our evaluation because it is public and often employed in empirical exercises similar to ours, such as \cite{wyc2015}, \cite{slsw2016}, \cite{zzwx2016} and \cite{TunuguntlaHoban21}. We direct the reader to \cite{zyws2015} for a detailed description of this data set.

A challenge we faced was that we could not find an RTB data set that recorded outcomes for advertisers when they lost auctions, that is, a data set that recorded $Y(0)$. We believe this further highlights a contribution of this paper: emphasizing non-zero payoffs for advertisers even when they lose an auction, as can be seen in equations (\ref{eq:revenueSPA}) and (\ref{eq:revenueFPA}). Despite being natural and critical from a causal inference perspective, accounting for $Y(0)$ in our context seems to have been widely ignored.

For our empirical evaluation, we use these data to calibrate the DGP, which is the same for SPAs and FPAs, as follows. We choose the milk powder advertiser and ad exchange number 3. We use mutually exclusive indicators for ten cities from the resulting sample as contexts, yielding $P=10$ different contexts, and set these contexts to be equiprobable. We use these probabilities to draw observations of each context on each round of this exercise.

The parameters of the distribution of $B_{CP}$ correspond to their MLE estimates. For potential outcomes, we specify a mixture of two lognormal distributions with mixing probability 0.55. The correlations between $\log Y(1)$ and $\log Y(0)$ are 0.3 and 0.5. Thus, even though our method treats potential outcomes as independent, under the true DGP they are correlated. The $ATE$s and the expectations of $Y(0)$ are random draws from a uniform distribution between 0 and 4, with the expectations of $Y(1)$ computed from these objects. The $\sigma^2$s of the first (second) component are random draws from a uniform distribution between 1.5 and 2 (0 and 0.5). We assume that the expectations of $Y(1)$ and $Y(0)$ are equal for the two components and solve for the $\delta$s to match these expectations.

In our simulations, we consider $E=200$ epochs. Within each epoch, we run the algorithm for $T=100$ rounds, and each round has $n=200$ observations. When running the MCMC methods, we set $Q=1,000$.

\begin{subsection}{Approaches under consideration}\label{sec:consider}

In addition to BITS, we implement in our simulations additional approaches to evaluate its performance related to both the inference and economic goals. We now briefly outline the other methods we consider. The first three methods rely on a grid with $r=1,\dots,R$ different bids; importantly, this grid always includes the true optimal bid. We provide a more detailed description of the implementation of these alternative approaches in Appendix \ref{app:alter}.

\paragraph{A/B test (AB)} \quad We implement an A/B test by randomizing with equal probability the bid placed from the grid. The $ATE$s are estimated by running a regression of $Y$ on $D$ separately for each $x$ using the available data. Under pure bid randomization, the estimated slope coefficients from these regressions are consistent for the $ATE$s due to Assumption \ref{ass:pv}. 

\paragraph{Explore-then-commit (ETC)} \quad The ETC approach proceeds as an A/B test for the first half of the experiment. It then collects all data from this half and, for each $x$, runs a regression of $Y$ on $D$ to estimate the $ATE$s. For SPAs, this approach places this estimate as the bid for arriving impressions, thus committing to what was learned in the first half of the experiment and to exploitation of that information in the second half. For FPAs, because the $ATE$ and $b^*$ are no longer equal, as Proposition \ref{prop:alignFPA} demonstrates, this approach computes the sample mean reward associated with each bid and submits the bid with the highest such mean for the second half of the experiment, akin to a greedy algorithm.

\paragraph{Off-the-shelf Thompson Sampling (TS\_OTS)} \quad In this off-the-shelf version of TS, rewards are assumed to follow independent normal distributions across arms for each $x$. The mean and variance of the distribution of arm $r$ are updated in the usual way using only observations obtained from pulling arm $r$. For each $x$, it places the estimated optimal bid. 

\paragraph{Correlated Thompson Sampling (TS\_CORR)} \quad This approach uses a reduced-form model for correlation in rewards across arms by specifying that rewards follow a normal distribution whose mean is a cubic polynomial of the bid. The parameters are updated as usual, and the method submits, for each value of the context $x$, the estimated optimal bid. 

\paragraph{Naive Bidding Thompson Sampling (BITS\_NAIVE)} \quad This version of BITS specifies a single lognormal distribution for potential outcomes and for $B_{CP}$, which facilitates implementation. Optimal bids are computed analogously to the procedure given above. 

We do not consider greedy algorithms in these simulations because our goal is not only to manage experimentation costs but also to perform inference on the estimated $ATE$s. The TS methods we consider are Bayesian and output objects we can use to perform inference on the $ATE$s, whereas greedy methods do not. However, there are greedy algorithms that could be easily applicable in this setting. They can be interpreted as frequentist versions of the TS algorithms we consider, which we address in more detail in Appendix \ref{app:alter}.

\end{subsection}

\begin{subsection}{Criteria of comparison}

\noindent BITS addresses two goals: minimizing the costs of experimentation and estimating the $ATE$s. Hence, to compare its performance to the five aforementioned alternative methods, we use two criteria, which assess each of these two goals. We describe them below.

\paragraph{Regret} \quad To evaluate the performance of these methods in tackling the economic goal, the metric we use is regret. For method $\iota$, the regret for context $x$ at the end of round $T$ is $\text{Regret}_{\iota}(T,x)=T\times \bar{\pi} \left [b^*(x) \right ]  - \sum_{t=1}^T  \mathbb{E} \left \{ \bar{\pi} \left [b_{\iota,t}(x) \right ] \right \}$; we have slightly abused notation by incorporating the number of observations into the number of rounds. The subscript $\iota$ indicates that the bid placed by each method can be different, and the randomness in these bids are reflected by the expectation. Finally, total regret is obtained by integrating $\text{Regret}_{\iota}(T,x)$ with respect to $x$: $\text{Regret}_{\iota}(T)= \sum_{p=1}^P \Pr (x = x_p) \times \text{Regret}_{\iota}(T,x)$.

\paragraph{Mean squared error (MSE)} \quad The MSE represents the inference goal. We compute it by using the final estimates of the $ATE$s over all the $e=1,...,E$ epochs: for context $x$, we estimate the MSE from method $\iota$ using $\widehat{\text{MSE}}_{\iota}(x)=\frac{1}{E}\sum_{e=1}^E \left (\widehat{ATE}_{\iota,e} (x)- ATE(x) \right )^2$. The term $\widehat{ATE}_{\iota,e}(x)$ is the estimate of $ATE(x)$ obtained at the end of epoch $e$ by method $\iota$. 

\end{subsection}

\subsection{Results}\label{sec:results}

We present the results from our simulations separately for SPAs and FPAs. For ease of illustration, we only show results for two of the ten contexts and present the full sets of results in Appendix \ref{app:full_res} as they are largely qualitatively similar. 

Our main interest is not on simply comparing the performance of the different methods, but also assessing the role of imposing structure on the method used to estimate $ATE$s and learn optimal bids. On one end of the spectrum, we have a method with pure randomization and that completely ignores the economic structure of the problem, the A/B test; on the other end, we have BITS, which tackles the inference and economic goals concurrently while exploiting for the full structure of the data and bid optimization problems. The remaining methods are intermediates between these two: ETC alters the A/B test by accounting for regret in the second half of the experiment; TS\_OTS accounts for regret from the beginning but still ignores the structure of the problem; TS\_CORR adds an extra layer by allowing rewards to be correlated across arms, albeit in a reduced-form way; and BITS\_NAIVE explicitly accounts for this structure but using a simpler parametrization. Comparing the performance of these different approaches can help us understand what the most important features are to successfully tackle the economic and inference goals.

\subsubsection{Second-price auctions}\label{sec:results_spas}

We start by analyzing the performance of the different methods in tackling the economic goal. Figure \ref{fig:reg_ctxt_spa} displays the evolution of cumulative regret across rounds averaged across the $E$ epochs for ETC, TS\_OTS, TS\_CORR, BITS\_NAIVE, and BITS. We omit the A/B test from this plot as we know that, by construction, its regret is linear, with slope corresponding to that of ETC during the first half of the experiment.

\begin{figure}[h]
    \centering
    \begin{subfloat}[\scriptsize Context 4\label{fig:reg_ctxt_spa4}]
        {\includegraphics[width=0.45\textwidth]{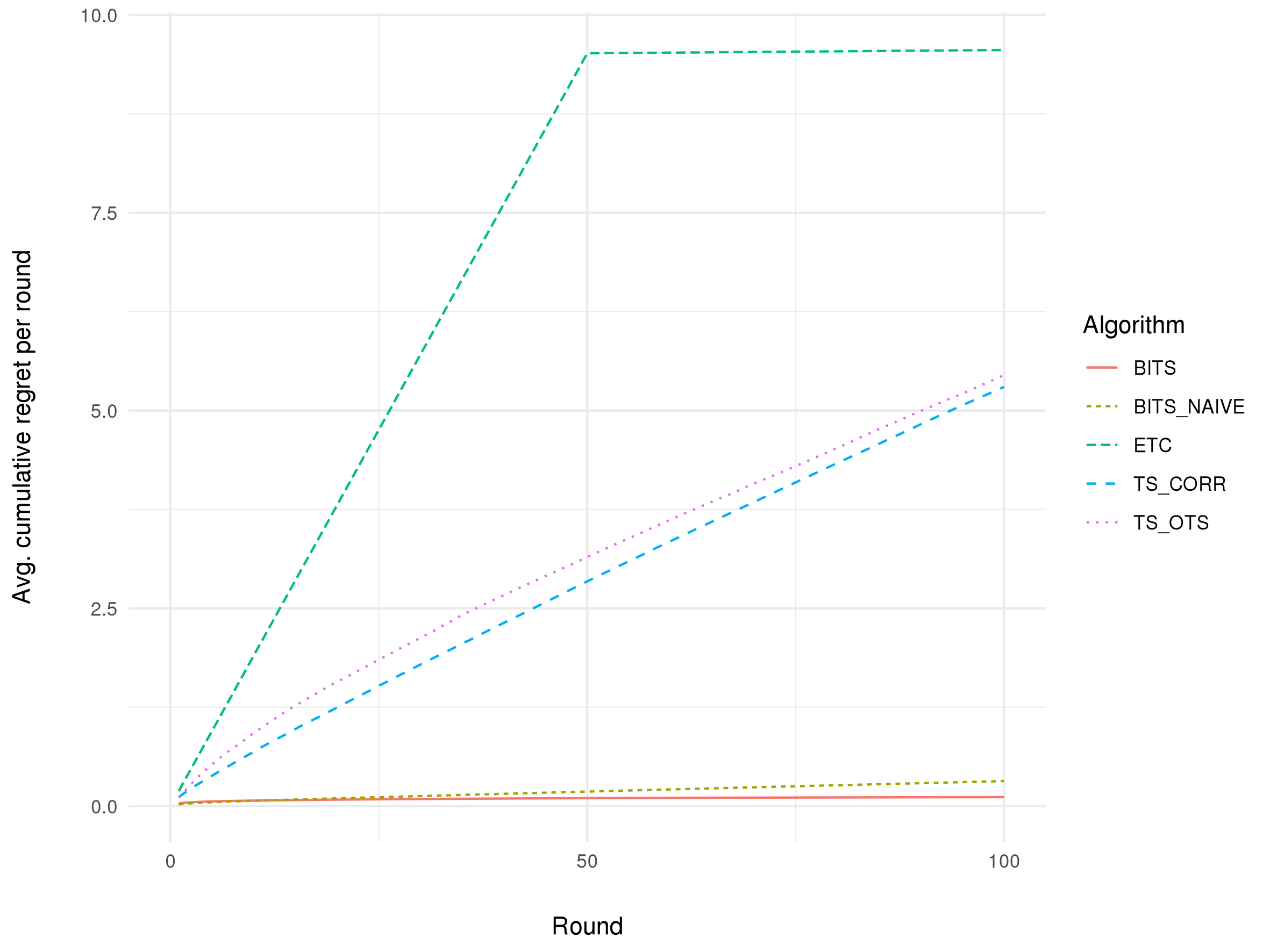}}
    \end{subfloat}
    \begin{subfloat}[\scriptsize Context 7\label{fig:reg_ctxt_spa7}]
        {\includegraphics[width=0.45\textwidth]{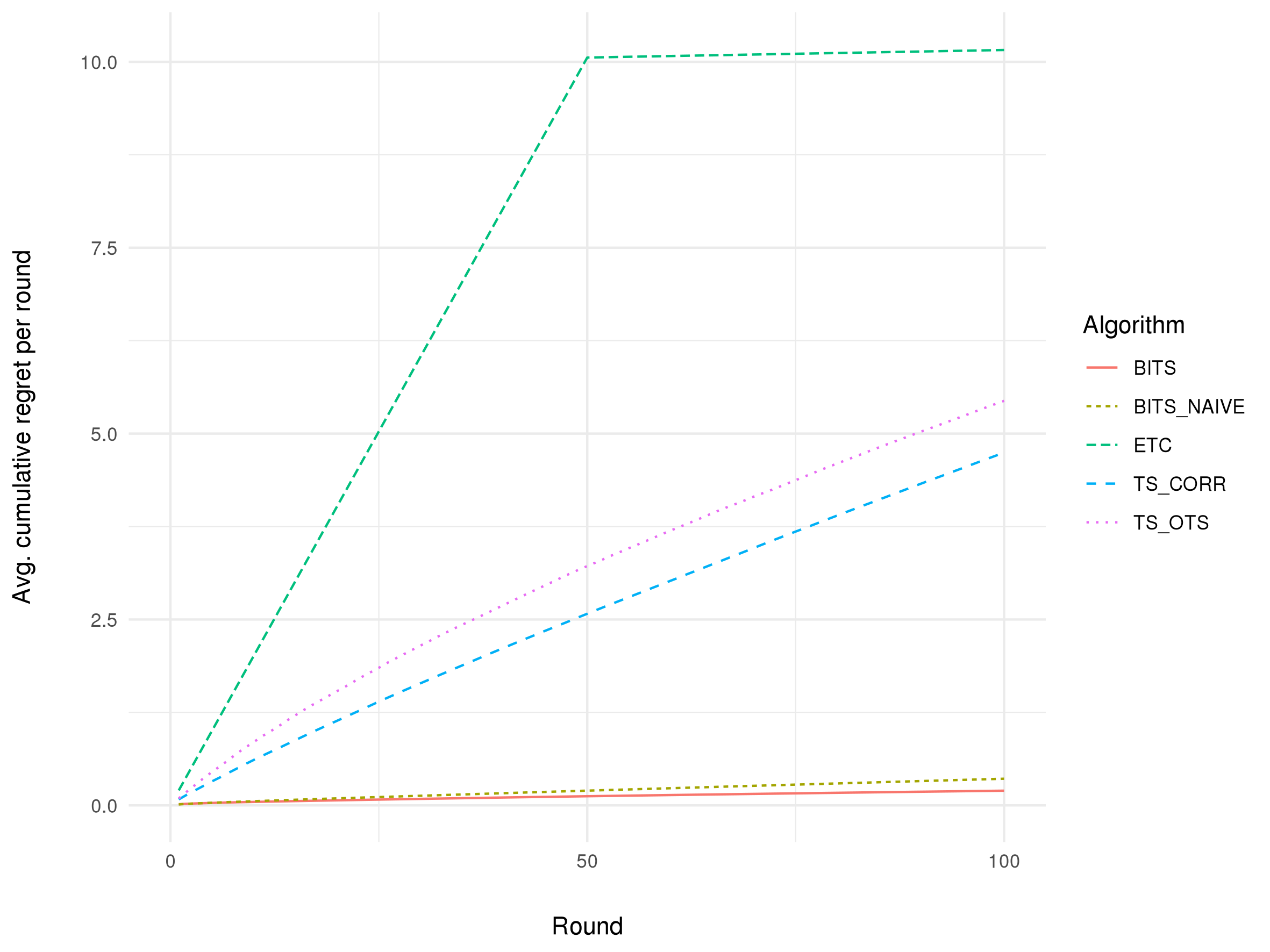}}
    \end{subfloat}\\
\caption{Evolution of cumulative regret for SPAs: contexts 4 and 7}
    \label{fig:reg_ctxt_spa}%
\end{figure}

Unsurprisingly, BITS has the best performance, being able to virtually eliminate regret as the experiment progresses, which demonstrates that it is able to correctly learn the optimal bid. On the other extreme, and also unexpectedly, ETC's regret displays a piecewise linear pattern. After the first half of the experiment, the slope approximates zero because the data gathered until then are sufficient to obtain a very precise estimate of the $ATE$, which, due to Proposition \ref{prop:alignSPA}, equals the true optimal bid.

TS\_OTS is an intermediate case between ETC and BITS as it tries to handle regret from the beginning of the experiment but, like ETC, does not fully account for the structure of the data, and particularly the relationship between observed and potential outcomes with treatment. Expectedly, TS\_OTS performs significantly better than ETC in minimizing regret for the duration of the experiment. In addition, if we augment this method by allowing for correlation in rewards across bids in a reduced-form way, which TS\_CORR does, the performance further improves. This improvement, however, is relatively smaller, and varies across contexts, indicating that its scope is dependent upon the underlying DGP.

Finally, we consider an approach that accounts for the structure of the data but uses a simpler model that does not match the real DGP, BITS\_NAIVE. Interestingly, this method approaches the performance of BITS very closely although it is in a sense misspecified. Its convergence shows that it is also able to recover the true optimal bid, which, in this case, coincides with the $ATE$. 

This is an important finding for two reasons. First, it suggests that using a simpler model, which is easier to implement and faster, may suffice to accomplish the advertiser's goals. Second, it highlights that allowing for correlation across arms, as TS\_CORR does, is not as important as accounting for the structure of the data for the purposes of minimizing regret.

We now proceed to analyze the performance of these methods in tackling the inference goal, that is, estimating the $ATE$. Figure \ref{fig:mse_ctxt_spa} shows the MSEs associated with the each method for the same contexts shown in Figure \ref{fig:reg_ctxt_spa}.

\begin{figure}[h]
    \centering
    \begin{subfloat}[\scriptsize Context 4\label{fig:mse_spa_ctxt4}]
        {\includegraphics[page=4, width=0.45\textwidth]{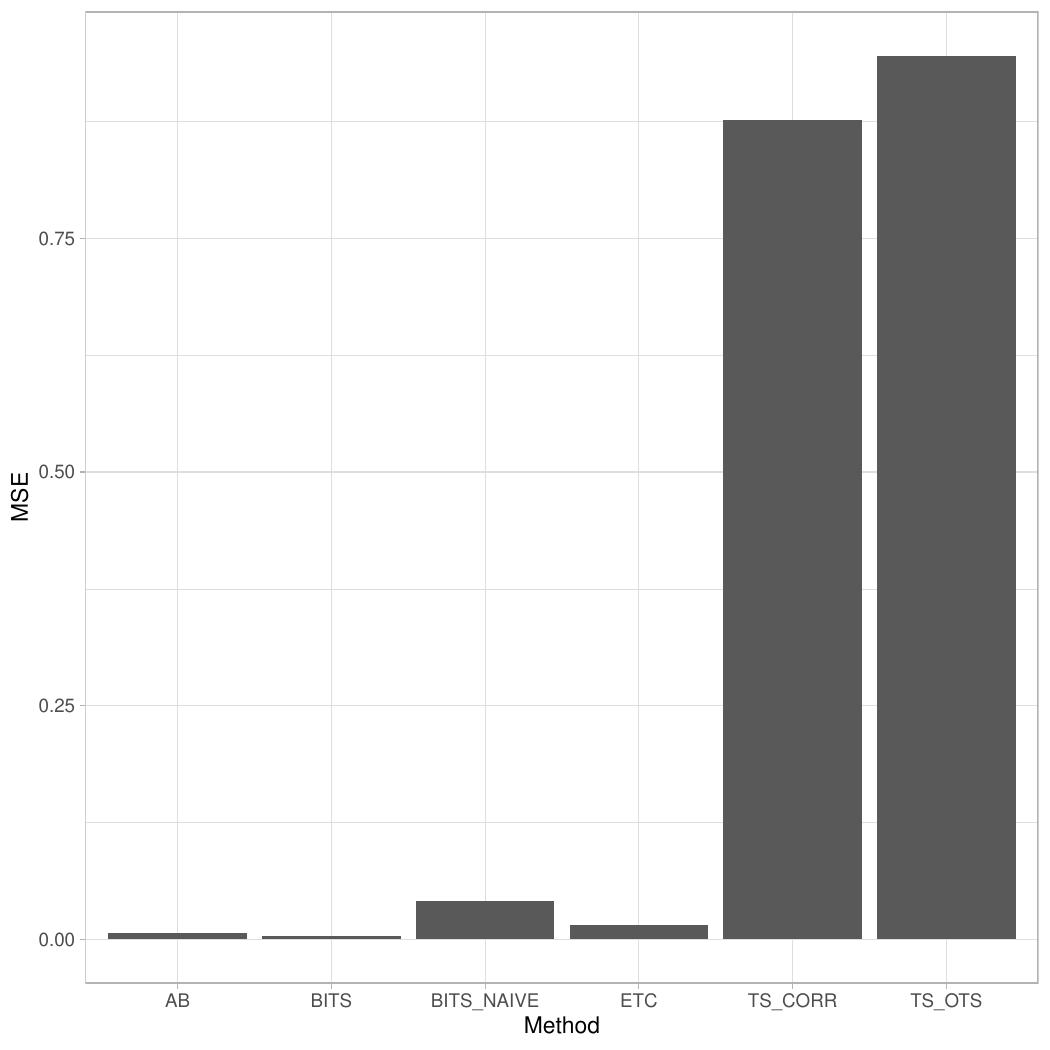}}
    \end{subfloat}
    \begin{subfloat}[\scriptsize Context 7\label{fig:mse_spa_ctxt7}]
        {\includegraphics[page=7, width=0.45\textwidth]{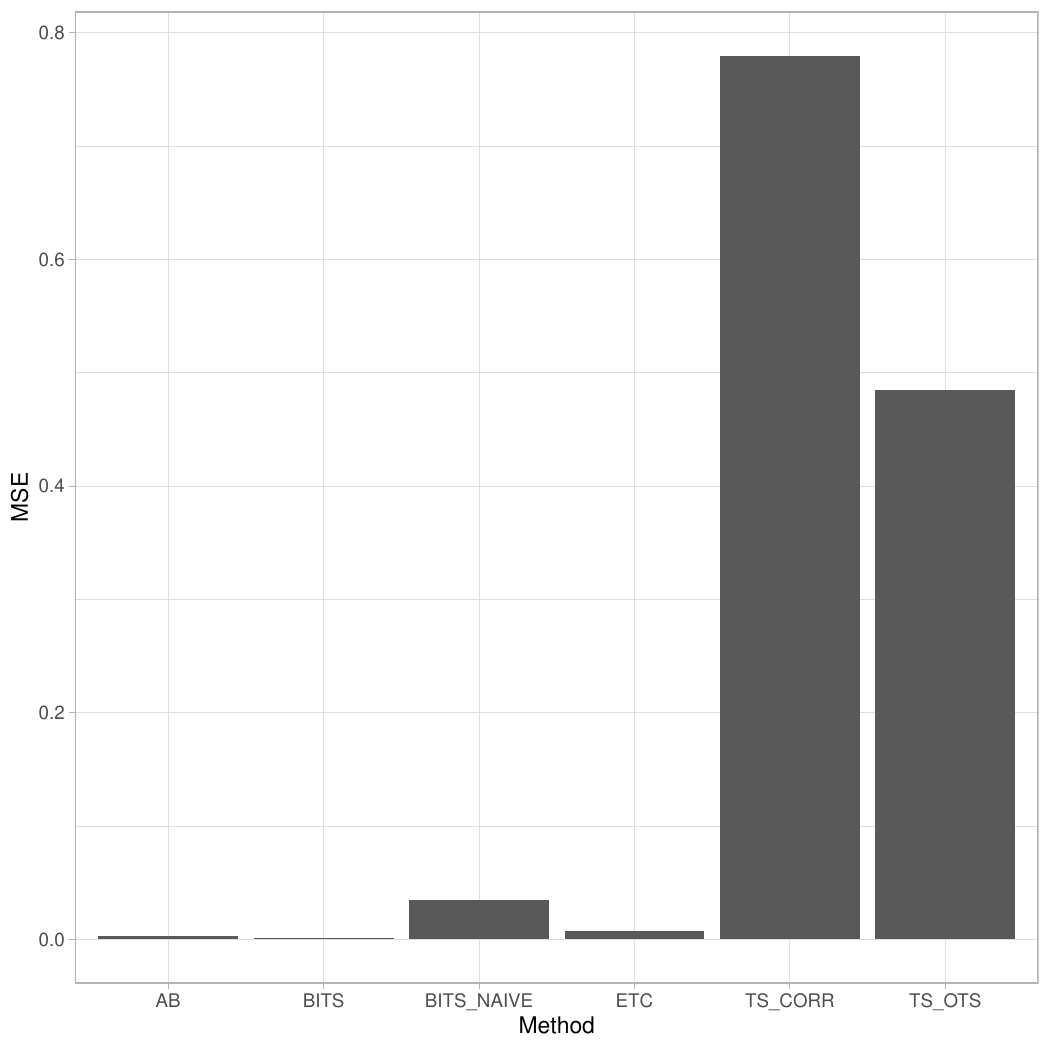}}
    \end{subfloat}\\
\caption{MSEs for SPAs: contexts 4 and 7}
    \label{fig:mse_ctxt_spa}%
\end{figure}

The results are intuitive. The A/B test achieves great performance in obtaining a low MSE as its estimate of the $ATE$ is consistent. The ETC also performs well but worse than the A/B test, which arguably occurs by construction since it uses the same estimator but with fewer observations. 

The MSE of BITS is virtually equal to that of the A/B test, albeit slightly smaller. This is unsurprising given that BITS imposes the correct parametric specification of the $ATE$ when estimating it, whereas the A/B test uses the OLS estimate.

In turn, the MSEs of TS\_OTS, TS\_CORR, and BITS\_NAIVE are somewhat aligned with their respective regrets. Interestingly, the ordering of the MSEs of TS\_OTS and TS\_CORR depends on the context. Furthermore, even though Figure \ref{fig:reg_ctxt_spa} suggests that these two methods can recover the true optimal bids with sufficient data and thus the $ATE$s, the precision of these estimates may be low.

The naive application of BITS also obtains a comparatively low MSE, although not as low as the more complex application of BITS, the A/B test approach, and ETC. Compounded with the findings of Figure \ref{fig:reg_ctxt_spa}, this suggests that the estimates from BITS\_NAIVE may not be as precise as those of these alternative methods.

\subsubsection{First-price auctions}\label{sec:results_fpas}

We now perform the same analysis for FPAs. This setting is in a sense more complex because $b^*$ and $ATE$ are no longer equal, as Proposition \ref{prop:alignFPA} shows, thus requiring a more involved estimator of the latter.

To assess the performance of these methods in tackling the advertiser's economic goal, we consider the evolution of cumulative regret across rounds averaged across the $E$ epochs. Figure \ref{fig:reg_ctxt_fpa} shows results analogous to those from Figure \ref{fig:reg_ctxt_spa}. Overall, these results are virtually equivalent to those obtained in a setting with SPAs. 

\begin{figure}[h]
    \centering
    \begin{subfloat}[\scriptsize Context 4\label{fig:reg_ctxt_spa4}]
        {\includegraphics[width=0.45\textwidth]{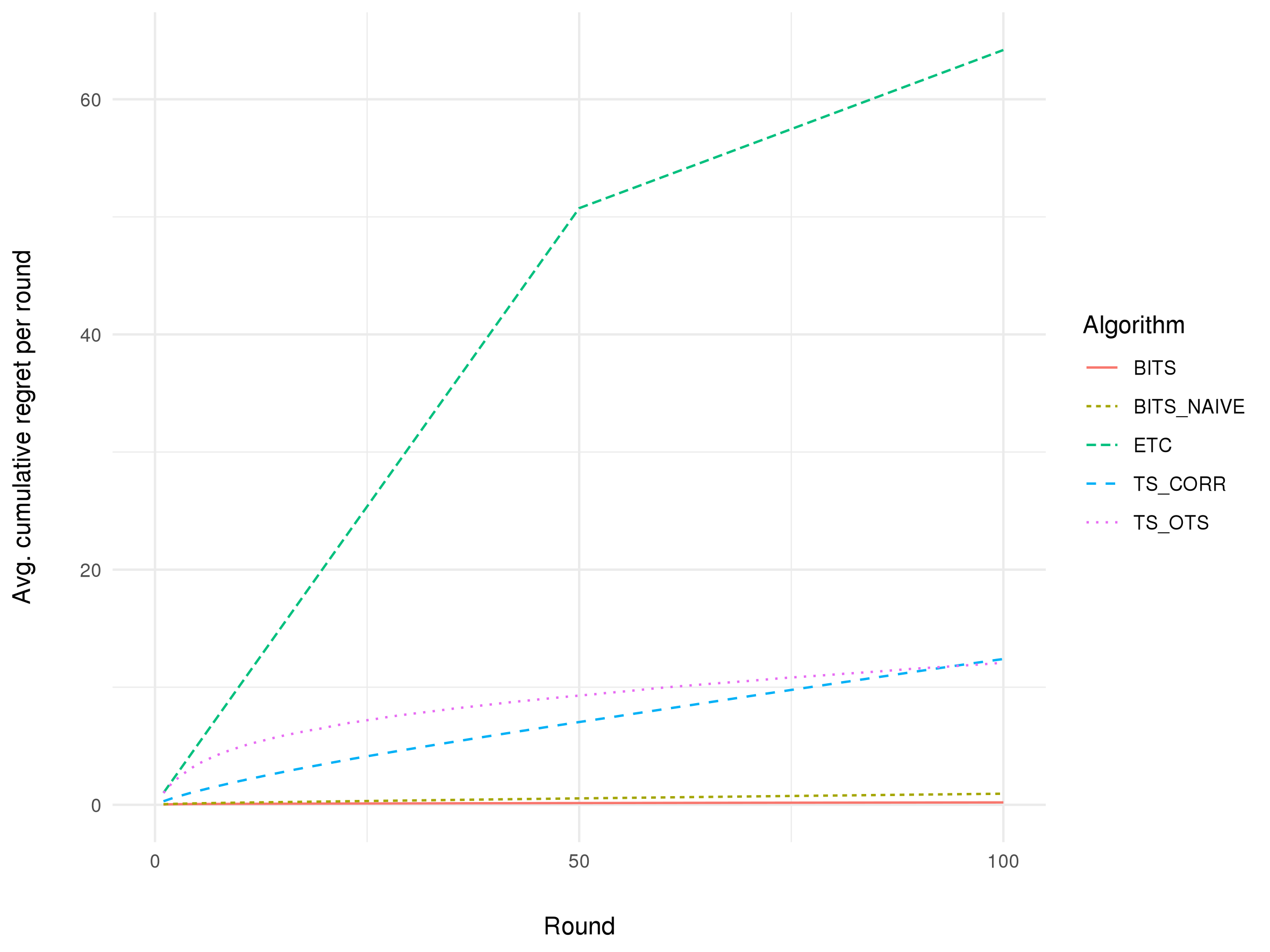}}
    \end{subfloat}
    \begin{subfloat}[\scriptsize Context 7\label{fig:reg_ctxt_spa4}]
        {\includegraphics[width=0.45\textwidth]{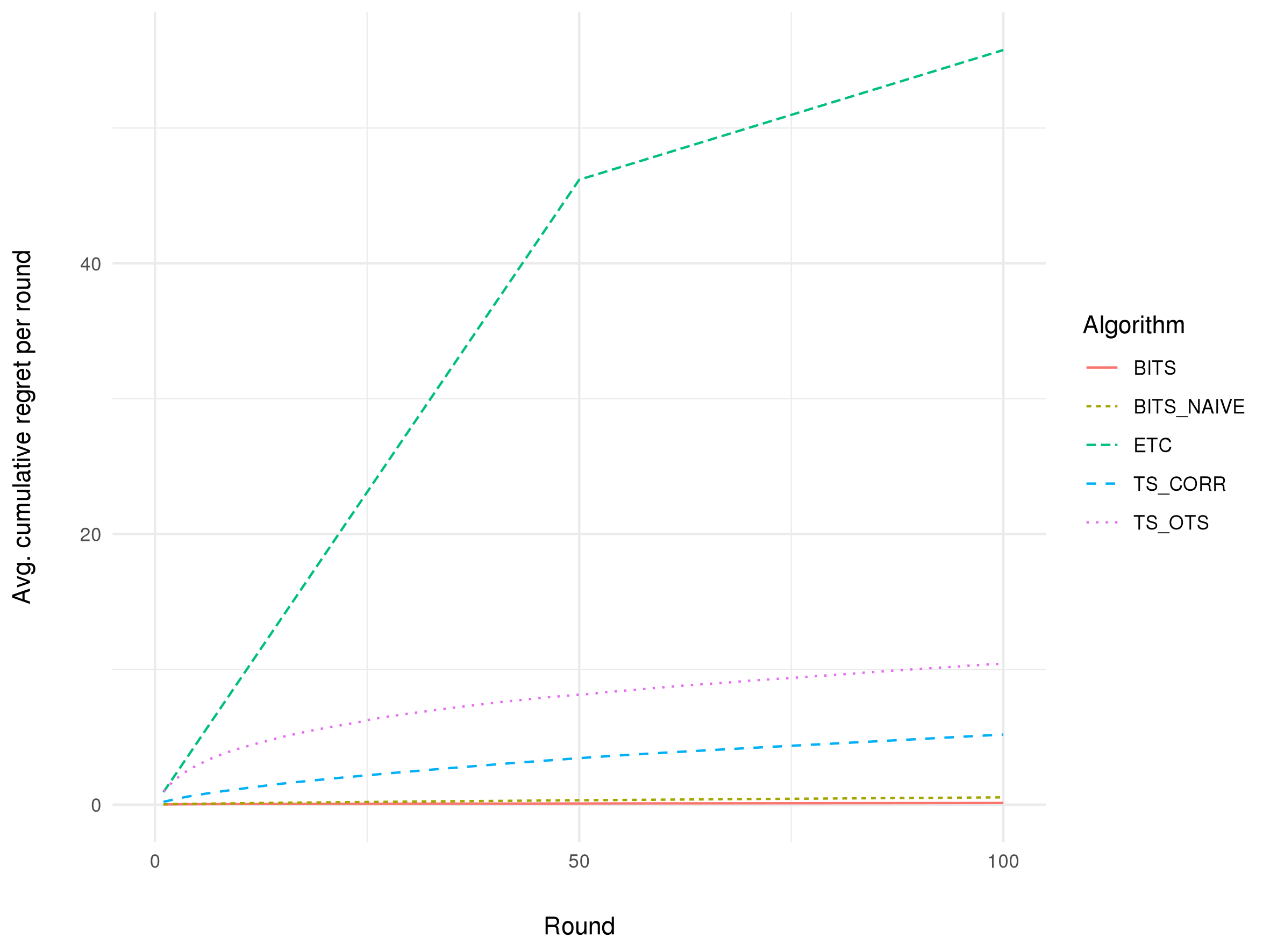}}
    \end{subfloat}\\
\caption{Evolution of cumulative regret for FPAs: contexts 4 and 7}
    \label{fig:reg_ctxt_fpa}%
\end{figure}

BITS has the best performance in minimizing regret, but BITS\_NAIVE performs similarly well. This is a notable result because this method in a sense misspecifies the underlying DGP in its implementation. As a consequence, it uses the incorrect distribution of $B_{CP}$ when calculating the bid adjustment factor characterized in Proposition \ref{prop:alignFPA}. Nevertheless, this discrepancy is effectively inconsequential.

The alternative adaptive approaches, TS\_OTS and TS\_CORR, can also correctly identify the optimal bid asymptotically However, allowing for correlation of rewards across arms in a reduced-form fashion may but need not improve their performance depending on the underlying DGP. In turn, ETC displays the same piecewise linear pattern as before. Nevertheless, the decrease in slope is not as pronounced as before.

We assess the performance of these methods in estimating the $ATE$s through MSEs, which Figure \ref{fig:mse_ctxt_fpa} presents analogously to from Figure \ref{fig:mse_ctxt_spa}, even though the difference in the scale of MSEs between the two scenarios is noteworthy. However, the displayed MSEs are only those associated with BITS, BITS\_NAIVE, the A/B test, and ETC. As Proposition \ref{prop:alignFPA} shows, under FPAs the $ATE$s are no longer equal to $b^*$. Obtaining estimates of these treatment effect parameters requires us to estimate a bid adjustment factor that depends on the distribution of $B_{CP}$. Consequently, reduced-form methods that ignore the structure of the data, such as TS\_OLS and TS\_CORR, are unable to estimate the $ATE$s.

\begin{figure}[h]
    \centering
    \begin{subfloat}[\scriptsize Context 4\label{fig:mse_fpa_ctxt4}]
        {\includegraphics[page=4, width=0.45\textwidth]{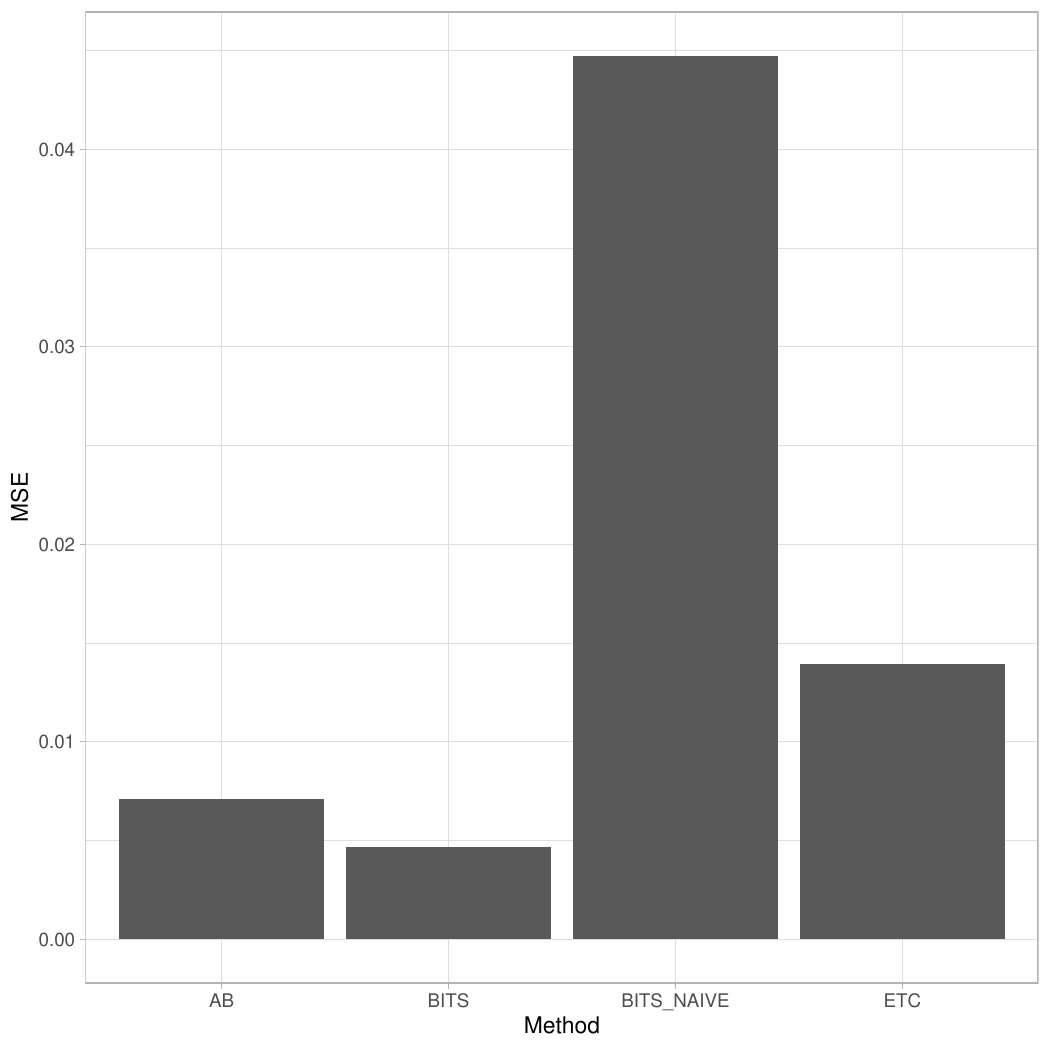}}
    \end{subfloat}
    \begin{subfloat}[\scriptsize Context 7\label{fig:mse_fpa_ctxt7}]
        {\includegraphics[page=7, width=0.45\textwidth]{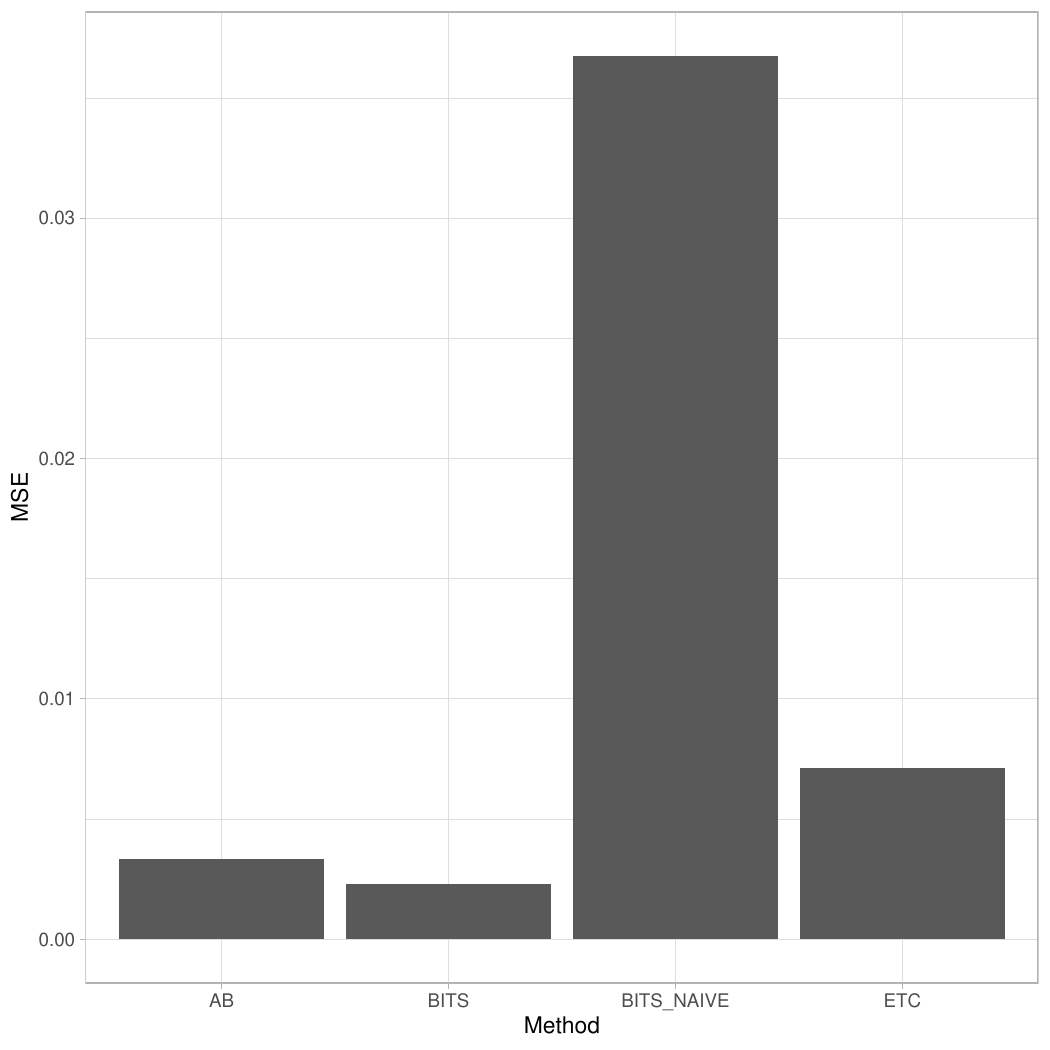}}
    \end{subfloat}\\
\caption{MSEs for FPAs: contexts 4 and 7}
    \label{fig:mse_ctxt_fpa}%
\end{figure}

As in the SPA setting, BITS performs best than the alternatives since it imposes the correct parametric form of the $ATE$. Its performance is followed by that of the A/B test, which, in turn, outperforms ETC by construction. Finally, the naive interpretation of BITS yields the highest MSE. Compounded with the results from Figure \ref{fig:reg_ctxt_fpa}, this suggests that despite being consistent, the estimator of the $ATE$ from BITS\_NAIVE may feature a lower precision than that of these alternative methods. 

\subsubsection{Summary}

Overall, the results from these exercises illustrate that BITS succeeds in addressing the advertiser's economic and inference goals. It is able to recover the true optimal bids under both SPAs and FPAs, which allows it to asymptotically eliminate regret, addressing the advertiser's economic goal. This result combined with the MSE of the estimator of the $ATE$s from BITS and Propositions \ref{prop:alignSPA} and \ref{prop:alignFPA} shows that this estimator is consistent, thus addressing the advertiser's inference goal.

Notably, the results also suggest that simpler implementations of BITS can also address these goals concurrently and successfully even if their models are in a sense misspecified, that is, they impose a parametric structure in their implementation that does not coincide with the true underlying DGP. This is a relevant result as their implementation can be substantially easier and faster than that of BITS. 

In addition, the naive implementation of BITS outperforms alternative adaptive methods, including an example that accounts for correlation in rewards across arms in a reduced-form way, which, in turn, outperforms simpler approaches that focus mostly on causal inference as opposed to managing regret. This suggests that exploiting the full structure of the data when addressing the advertiser's goals can be as relevant as accounting for such correlation, if not more.

Finally, it is important to notice that with additional data and assumptions other approaches can be expected to outperform BITS. This could be the case, for example, when one has access to data on clicks and sales, one could impose an explicit connection between displays and sales via clicks using the method and assumptions from \cite{bgh2021}. BITS is agnostic about such connections and would not exploit such data if they were available. We expect that if the structure a method imposes on these connections is correct, then such method would naturally perform better than BITS. 

\end{section}

\begin{section}{Concluding remarks}

This paper introduced an online method to perform causal inference on RTB advertising. This method leverages the alignment between the optimal bids under sealed-bid SPAs and FPAs with the expected effect of ad exposure to concurrently estimate such effects while accounting for the costs of experimentation. The specific mappings between optimal bids and the expected effect of ad exposure are derived by leveraging auction theory and by carefully outlining the conditions under which they hold.

Some broader implications of the experimental design beyond RTB advertising are worth mentioning. First, the ideas presented here can be useful in other situations outside of ad auctions where there is a cost to obtain experimental units and where managing these costs is critical for the viability of the experiment. These costs can be driven by the cost of acquiring experimental units (such as sending mailers in offline experiments or other incentives given for experimental participation), as well as the opportunity costs of showing unproven treatments (such as new creatives or new price ranges) to users. 

Another implication is that, in business experiments, incorporating the firm's payoff or profit maximization goal into the allocation and acquisition of experimental units is helpful. Examples include pricing experiments and product feature experiments for which there is a clear payoff goal to the firm from such experimentation, which can be incorporated into a MAB problem. Given the burgeoning utilization of experimentation by firms, we believe that leveraging this perspective in business experiments more broadly has value. 

Finally, another takeaway from the proposed approach is that it demonstrates the value of embedding experimental design in the microfoundations of the problem, which enables leveraging economic theory to make progress on running experiments more effectively. This aspect could be utilized in other settings where large-scale experimentation is feasible and where economic theory puts structure on the behavior of agents and associated outcomes. The fusing of experimental design and economic theory in this manner is a direction for our future research.

\end{section}

{\footnotesize\bibliography{external_ref2}}

\newpage

\appendix
\renewcommand{\thesubsection}{\Alph{subsection}}
\counterwithin{table}{subsection}   
\counterwithin{figure}{subsection} 
\numberwithin{equation}{subsection}

\section*{Appendix}

\subsection{Proofs}\label{app:proofs}

\subsubsection{Proof or Proposition \ref{prop:alignSPA}}

\begin{proof}{
To prove Proposition \ref{prop:alignSPA}, we first rewrite equation (\ref{eq:auc_epaySPA}):
\begin{align*}
	  \bar{\pi}(b\vert x)  
	  &= \Pr \left \{ B_{CP} \leq b \middle\vert x \right \} \left \{ \mathbb{E} \left [ Y(1) -Y(0) \middle \vert B_{CP} \leq b ;x  \right ] - \mathbb{E} \left [ B_{CP} \middle \vert B_{CP} \leq b ;x \right ] \right \}  \\
	  &= \Pr \left \{ B_{CP} \leq b \middle\vert x \right \} \left \{ \mathbb{E} \left [ Y(1) -Y(0) \middle \vert x  \right ] - \mathbb{E} \left [ B_{CP} \middle \vert B_{CP} \leq b ;x \right ] \right \}  \\
	  	  &= \Pr \left \{ B_{CP} \leq b \middle\vert x \right \} \left \{ ATE(x) - \mathbb{E} \left [ B_{CP} \middle \vert B_{CP} \leq b;x \right ] \right \}  \\
	  &=\int_0^b \left [ATE(x) - b_{CP} \right ]f_{CP}(b_{CP}|x)db_{CP} ,
\end{align*}
where the second equality follows from Assumption \ref{ass:pv}. Since the advertiser cannot submit negative bids, if $ATE(x)\leq 0$ then $b^*(x)=0$ because the integrand is strictly negative. Otherwise, notice that the integrand is non-negative as long as $b(x) \leq ATE(x)$, which implies the optimal bid cannot exceed $ATE(x)$. Due to Assumption \ref{ass:pos_dens}, $f_{CP}(\cdot|x)>0$ around $ATE(x)$ so that $b^*(x)=ATE(x)$.}
\end{proof}

\subsubsection{Proof or Proposition \ref{prop:alignFPA}}

\begin{proof}{
To prove Proposition \ref{prop:alignFPA}, we proceed as above by first rewriting equation (\ref{eq:auc_epayFPA}):
\begin{align*}
\bar{\pi}(b\vert x)&=\Pr\left\{ B_{CP}\leq b\middle\vert x\right\} \left\{ \mathbb{E}\left[Y(1)-Y(0)-b\middle\vert B_{CP}\leq b;x\right]\right\} \\&=\Pr\left\{ B_{CP}\leq b\middle\vert x\right\} \left\{ \mathbb{E}\left[Y(1)-Y(0)\middle\vert x\right]-b\right\} \\&=F_{CP}\left(b|x\right)\left\{ ATE(x)-b\right\}, 
\end{align*}
where, once again, the second equality follows from Assumption \ref{ass:pv}. If $ATE(x)\leq 0$, it is straightforward to verify that $b^*(x)=0$ since the advertiser cannot submit negative bids. Consider now the case where $ATE(x)>0$. The first-order condition with respect to $b$ is:
\begin{align}\label{eq:ATE-FPA}
ATE(x)=b+\frac{F_{CP}\left(b|x\right)}{f_{CP}\left(b|x\right)}\equiv \chi (b).
\end{align}
Assumptions \ref{ass:pos_dens} and \ref{ass:rev_haz} imply that the right-hand side of equation (\ref{eq:ATE-FPA}) is monotonically increasing in $b$ around $ATE(x)$. Hence, $\chi(\cdot)$ is invertible, which yields a unique solution.\footnote{This demonstrates that Assumption \ref{ass:rev_haz} is just a sufficient condition to obtain uniqueness of $b^*(x)$. Invertibility of $\chi(b|x)$ holds provided that $\frac{F_{CP}\left(b|x\right)}{f_{CP}\left(b|x\right)}$ does not decrease faster than $b$ increases. Thus, under appropriate smoothness conditions Assumption \ref{ass:rev_haz} could be replaced by the assumption that $\frac{\frac{F_{CP}\left(b|x\right)}{f_{CP}\left(b|x\right)}}{\partial b}>-1$.} Denoting its inverse by $\chi^{-1}(\cdot)$, the optimal bid is therefore given by $b^*(x) = \max \left\{0,\chi^{-1}\left [ATE(x) \right ] \right \}$.}
\end{proof}

\subsection{Drawing from posterior distributions: MCMC methods}\label{App:full-cond-gibbs}

\noindent We now present in more detail the procedure we summarized in Section \ref{sssec:post_summ} that obtains draws from the posterior distribution of the parameters given the data. First, we describe the procedure to obtain draws of the parameters of the distributions of potential outcomes. Then, we describe the procedure for those of the distribution of the highest competing bid. These two procedures can be conducted separately because of Assumption \ref{ass:pv}.

\subsubsection{Potential outcomes: Gibbs sampling with data augmentation}\label{app:potout}

As we mentioned in Section \ref{sssec:post_summ}, we obtain draws of the parameters associated with $Y_(1)$, given by $\theta_{Y,1} = \left [\delta_{k,1}, \sigma_{k,1}^2, \xi_{k,1} \right ]_{k=1}^{K}$, separately from those of $Y(0)$, $\theta_{Y,0} = \left [\delta_{k,0}, \sigma_{k,0}^2, \xi_{k,0} \right ]_{k=1}^{K}$. Hence, we describe the procedure regarding the vector $\theta_{Y,d}$, where $d\in\{0,1\}$.

\paragraph{Priors} \quad For the sake of convenience, we choose the following prior distributions:
\begin{align}\label{eq:priors}
    \begin{split}
        &\delta_{k,d}\sim N \left (\mu_{k,d}, V_{k,d}^{-1} \right ) \\
        &\sigma_{k,d}^{-2}\sim \Gamma \left (\alpha_{k,d}, \beta_{k,d} \right ) \\
        &\xi_{d}\sim \mathcal{D} \left (\aleph_d \right )
    \end{split}
\end{align}
where: the $\mu$s are $P$-by-1 vectors; all $V$s are positive semidefinite matrices of size $P$; $\Gamma(\alpha,\beta)$ denotes the gamma distribution with shape parameter $\alpha$ and rate parameter $\beta$, with $\alpha_{k,d}$ and $\beta_{k,d}$ being non-negative for all $k$ and $d$; $\xi_{d}\equiv \left [\xi_{1,d}, \cdots, \xi_{K_Y,d} \right ]'$; $\mathcal{D}(\cdot)$ denotes the Dirichlet distribution;  and $\aleph_{d}\equiv \left [\aleph_{1,d}, \cdots, \aleph_{K_Y,d} \right ]'$, with $\aleph_{k,d}>0$ for all $k$ and $d$.

Following the typical approach for Bayesian estimation of finite mixture models, we augment the data with component identifiers. Denote observation $i$'s vector of identifiers by $z_{id}$, with $z_{ikd}$ being the $k$-th element of the vector, and the initial value of $\xi_d$ by $\xi_d^{(0)}$. The prior for $z_{id}$ is: 
\begin{align}\label{eq:prior_z}
    z_{id}\overset{i.i.d.}\sim \text{Multinomial}\left(\xi_d^{(0)}\right),
\end{align}
which is also chosen solely for the sake of convenience. 

\paragraph{Augmenting potential outcomes} \quad Assume the algorithm is in round $q$, with previous draws of the relevant parameters identified by the superscript $(q-1)$. The first step in the algorithm is to draw the missing values of the potential outcomes so that they can be augmented. Given the distributional assumptions, the priors, and the previous draws of parameters, it follows that the missing value, $Y_i^m$, follows the following distribution:
\begin{align}\label{eq:missY}
    Y_i^{m}| Y_i,X_i,D_i=d,z_{ikd}^{(q-1)}=1; \theta_Y^{(q-1)} \sim N \left (X_i' \delta_{k,d}^{(q-1)}, \sigma_{k,d}^{2,(q-1)} \right ),
\end{align}
which is straightforward to draw from. Let this draw be given by $Y_i^{m,(q)}$. Then the augmented potential outcomes are:
\begin{align}\label{eq:augY}
    \begin{split}
        \tilde{Y}_{i}^{(q)}(1)&= D_i Y_i + (1-D_i)Y_i^{m,(q)}  \\
        \tilde{Y}_{i}^{(q)}(0)&= D_iY_i^{m,(q)} + (1-D_i) Y_i
    \end{split}.
\end{align}

\paragraph{Drawing new normal distributions parameters} \quad The next step is to draw new values of $\delta$s and $\sigma^2$s. To do so, we first establish some notation.

Define $\tilde{Y}^{(q)}(d)$ as the $n$-by-1 vector that stacks the values $\tilde{Y}_{i}^{(q)}(d)$ and $X$ be the $n$-by-$P$ matrix that stacks the $X_i'$ vectors across all $i$. Let $n_{k,d}=\sum_{i=1}^n \mathbbm {1} \left \{z_{ikd}=1 \right \}$. Define $\tilde{Y}_k^{(q)}(d)$ as the $n_k$-by-1 vector that stacks the values $\tilde{Y}^{(q)}(d)$ only for the observations such that $z_{ikd}=1$ and define $X_k$ and $z_{kd}$ analogously. Given our assumptions, it follows that:
\begin{align}\label{eq:postNsig}
        &\sigma_{k,d}^{2,(q)}|\tilde{Y}^{(q)}(d),X,D=d,z_{kd}=1;\delta_{k,d}^{(q-1)} \\
        & \qquad \sim \Gamma \left (\alpha_{k,d} + \frac{n_{kd}}{2}, \beta_{k,d} + \frac{\left ( \tilde{Y}_k^{(q)}(d) - X_k \delta_{k,d}^{(q-1)} \right )'\left ( \tilde{Y}_k^{(q)}(d) - X_k \delta_{k,d}^{(q-1)} \right )}{2} \right ) \\
\end{align}
and
\begin{align}\label{eq:postNdel}
    &\delta_{k,d}^{(q)}|\tilde{Y}^{(q)}(d),X,D=d,z_{kd}=1; \sigma_{k,d}^{2,(q)} \\
        & \qquad \sim N \left ( \left [ \frac{X_k'X_k}{\sigma_{k,d}^{2,(q)}} + V_{k,d} \right ]^{-1} \left [ \frac{X_k'\tilde{Y}^{(q)}_k(d)}{\sigma_{k,d}^{2,(q)}} + V_{k,d}\mu_{k,d} \right ],\left [ \frac{X_k'X_k}{\sigma_{k,d}^{2,(q)}} + V_{k,d} \right ]^{-1} \right ).
\end{align}
Once again, it is straightforward to obtain draws from these distributions.

\paragraph{Drawing new component identifiers} \quad Given our assumptions, the posterior distribution of the component identifiers is:
\begin{align}\label{eq:postz}
    \begin{split}
        &z_{id}^{(q)}|\tilde{Y}^{(q)}(d),X,D=d;\delta_{d}^{(q)},\sigma_{d}^{2,(q)},\xi_d^{(q-1)} \\
        & \qquad \sim \text{Multinomial} \left ( \frac{\xi_{1,d}^{(q-1)}\phi \left ( \tilde{Y}_i^{(q)}(d);X_i\delta_{1,d}^{(q)}; \sigma_{1,d}^{2,(q)} \right )}{\sum_{k=1}^{K_Y}\xi_{k,d}^{(q-1)}\phi \left ( \tilde{Y}_i^{(q)}(d);X_i\delta_{k,d}^{(q)}; \sigma_{k,d}^{2,(q)} \right )}, \dots, \right . \\
        & \qquad \qquad \qquad \qquad \qquad \qquad \left .\dots , \frac{\xi_{K_Y,d}^{(q-1)}\phi  \left ( \tilde{Y}_i^{(q)}(d);X_i\delta_{K_Y,d}^{(q)}; \sigma_{K_Y,d}^{2,(q)} \right )}{\sum_{k=1}^{K_Y}\xi_{k,d}^{(q-1)}\phi \left ( \tilde{Y}_i^{(q)}(d);X_i\delta_{k,d}^{(q)}; \sigma_{k,d}^{2,(q)} \right )} \right ),
    \end{split}
\end{align}
where $\phi(Y;\delta;\sigma^2)$ denotes the standard normal density with mean $\delta$ and variance $\sigma^2$ evaluated at $Y$. Thus, it is straightforward to draw new component identifiers.

\paragraph{Drawing new mixing probabilities} \quad The last step is to draw new mixing probabilities. This is trivial because, given our assumptions, their posterior distribution is:
\begin{align}\label{eq:postxi}
    \xi_d^{(q)}|z_d \sim \mathcal{D}\left (\aleph_{1,d} + n_{1,d}, \dots, \aleph_{K_Y,d} + n_{K_Y,d} \right ).
\end{align}

\paragraph{Summary} \quad To summarize, the procedure starts with values for the parameters of the prior distributions and initial values for the parameters. By cycling through equations (\ref{eq:missY})$-$(\ref{eq:postxi}), the researcher can obtain draws from the posterior distributions of the parameters $\theta_Y$ given the data.

\subsubsection{Highest competing bid: random walk Metropolis-Hastings}

We implement a standard random walk Metropolis-Hastings procedure using a multivariate normal proposal distribution. To be precise, let $\mathcal{L} \left (D,X,b \middle \vert \delta_{CP} \right )$ be the likelihood of the elements of the data associated with $B_{CP}$.

The procedure works as follows:
\begin{enumerate}
    
    \item Initiate with a random value $\delta_{CP}^{(0)}$.

    \item For $q=1,\dots,Q$:

    \begin{itemize}
        
        \item[a.] Draw $\delta_{CP}^*$ from $N\left (\delta_{CP}^{(q)}, V_{CP}  \right )$.

        \item[b.] Define:
        \begin{align}\label{eq:prob_mh}
            \zeta=\min \left \{ \frac{\mathcal{L} \left (D,X,b \middle \vert \delta_{CP}^* \right )}{\mathcal{L} \left (D,X,b \middle \vert \delta_{CP}^{(q)} \right )},1  \right \}.
        \end{align}

        \item[c.] Take:
        \begin{align}\label{eq:new_draw_mh}
            \delta_{CP}^{(q+1)}= \begin{cases} \delta_{CP}^* & \text{with probability } \zeta \\ \delta_{CP}^{(q)} & \text{with probability } 1-\zeta \end{cases}.
        \end{align}
    
    \end{itemize}

\end{enumerate}

\subsection{Additional considerations}\label{app:add}

\noindent We now discuss some practical considerations that arise when implementing the algorithm and ways in which it can be adapted or extended to accommodate variations in the experimentation environment and advertiser goals.

\subsubsection{Regret minimization versus best arm identification}

\noindent We implement BITS under a regret minimization framework based on the viewpoint that the advertiser seeks to maximize payoffs from auction participation during the experiment. We follow this approach because our setting involves monetary payments. However, we could alternatively cast the problem as one of best arm identification as in \cite{Bubecketal2009}, \cite{GKL2016} and \cite{DNCP2019}. Under this formulation, the problem is cast in terms of pure exploration, so the role of adaptive experimentation is to obtain information before committing to a final decision involving the best arm identified with that information.\footnote{\cite{russo16} provides an adaptation of TS to best arm identification.} To leverage Propositions \ref{prop:alignSPA} and \ref{prop:alignFPA}, all we need is a MAB framework to recover the bid with highest expected reward, so the core idea behind our proposed approach ports in a straightforward way to this alternative formulation of the advertiser's objectives.

\subsubsection{Alternative parametric assumptions}

More flexible parametric specifications may be used instead of the one we have proposed. One potential concern is the context vector $X$, which can be high-dimensional. If so, it is possible to alter the prior distributions we considered to perform additional regularization and potentially variable selection.

A different concern regards the choice of distribution for the highest competing bid, $B_{CP}$. This is especially the case under FPAs because it directly impacts the optimal bids. Even though our choice of a Fr\'{e}chet distribution is motivated by theory, it can still be considered too restrictive. Unfortunately, data restrictions under FPAs, namely the fact that $B_{CP}$ is never observed, hinder the use of more flexible parametrizations. However, should $B_{CP}$ become observable due to information sharing by the AdX, for example, then the same mixture of distributions model could be used to model its distribution.

\subsubsection{Budget constraints}

\noindent The current formulation does not explicitly incorporate budget constraints, which can be relevant in practice. While there are more general methods to perform bid optimization in the presence of budget constraints, such as \cite{cai2017real} and \cite{TunuguntlaHoban21}, these methods do not focus on performing causal inference. Part of the complication is that the optimal bidding policy becomes a function both of $x$ and of the remaining budget and linking it to $ATE$s becomes non-trivial. Formal incorporation of budget constraints would therefore need extending the algorithm beyond its current scope and is left as a topic for future research. From a practical perspective, our method could also be implemented so that the advertiser learns $ATE(x)$, which, can then be inserted into an optimization algorithm that accounts for budget constraints explicitly.

\subsubsection{Repeated exposures}

Our method does not explicitly account for repeated exposures of ads to the same user. Although a partial fix to this limitation would be to include the number of exposures to the context $x$, a more complete solution would be to explicitly incorporate repeated exposures by borrowing ideas from \cite{bgh2021}, for example. We leave this exercise for future research.



\subsection{Alternative approaches}\label{app:alter}

We now describe in more detail the alternative methods to estimate $ATE$s from Section \ref{sec:consider}. To this end, we consider a grid of $R$ bids denoted by $b_r$ with $r=1,\dots,R$, which always contain the true optimal bid. All the methods are applied separately for each value of the context $x$ except for the naive implementation of BITS.

\paragraph{A/B test (AB)} \quad The A/B test proceeds as follows: 
\begin{enumerate}

    \item For each $i$, draw $b_r$ with probability $\frac{1}{R}$ and place it as the bid.
    
    \item Record $D_i$ and $Y_i$.
    
    \item At the end of the experiment, estimate the parameters of the regression equation $Y_i=\lambda_0 + (\lambda_1 - \lambda_0) D_i + \epsilon_i$ via OLS, where $i=1,\dots,n$.
    
    \item Set $\widehat{ATE}=\hat{\lambda}_1 - \hat{\lambda}_0$, where $\hat{\lambda}_1 - \hat{\lambda}_0$ is the OLS estimate of $ATE$.
    
\end{enumerate}

The A/B test does not attempt to minimize regret, which, consequently, is constant.

\paragraph{Explore-then-commit (ETC)} \quad The ETC procedure is as follows: 
\begin{enumerate}

    \item For each $i$  during the first half of the experiment, draw $b_r$ with probability $\frac{1}{R}$ and place it as the bid.
    
    \item Record $D_i$, $b_i$, and $Y_i$.
    
    \item At the end of the first half of the experiment, estimate the parameters of the regression equation $Y_i=\lambda_0 + (\lambda_1 - \lambda_0) D_i + \epsilon_i$ via OLS, where $i=1,\dots,\frac{n}{2}$.
    
    \item Set $\widehat{ATE}=\hat{\lambda}_1 - \hat{\lambda}_0$, where $\hat{\lambda}_1 - \hat{\lambda}_0$ is the OLS estimate of $ATE$.
    
    \item For the second half of the experiment:
    
    \begin{enumerate}
    
        \item Place $\max \left \{ \widehat{ATE},0 \right \}$ as the bid if the auction is an SPA.
        
        \item For each $r$, compute $\hat{\pi}(b_r)=\frac{\sum_{i=1}^n \mathbbm{1}\{b_i=b_r\} \left ( Y_i - D_i b_i\right )}{\sum_{i=1}^n \mathbbm{1}\{b_i=b_r\}}$ and place $\max \left \{ b_{r^*},0 \right \}$ as the bid, where $r^* = {\arg\max}_{r=1,\dots,R} \hat{\pi}(b_r) $ if the auction is an FPA.
        
    \end{enumerate}

\end{enumerate}

ETC's cumulative regret is piecewise linear: it coincides with that of the A/B test for the first half of the experiment and then is implied by the bid that is placed for the second half. We expect that the larger $n$ is, the ``flatter'' the linear regret is for the second half of the experiment as $\hat{\lambda}_1 - \hat{\lambda}_0$ should become closer to $ATE$. 

Notice that the bids that are placed during the second half of the experiment are computed differently depending on the auction format. Under SPAs, Proposition \ref{prop:alignSPA} implies that the optimal bid coincides with the $ATE$; thus, it can be leveraged by using $\widehat{ATE}$. Under FPAs, this equivalence no longer holds, and because ETC ignores the underlying structure of the problem, the estimated optimal bid is computed through the sample average of rewards.

\paragraph{Off-the-shelf Thompson Sampling (TS\_OTS)} \quad We implement TS\_OTS by setting a different and independent normal distribution for each arm $r$. More precisely:
\begin{enumerate}

    \item The likelihood is given by $\pi_i(b_r)\sim N(\delta_r,\sigma_r^2)$.
    
    \item The priors are $\delta_r \sim N(\mu_r,V_r^{-1})$ and $\sigma^{-2}_r\sim \Gamma(\alpha_r,\beta_r)$.
    
    \item For each $i$, after submitting bid $b_i$: 
    
    \begin{enumerate}
    
        \item Compute $\pi_i = Y_i - D_i B_{CP,i}$ if the auction is an SPA.

        \item Compute $\pi_i = Y_i - D_i b_i$ if the auction is an FPA.
        
    \end{enumerate}

    \item For each $r$, draw from the posterior distribution of $\delta_r$ and $\sigma_r^2$ using only observations such that $b_i=b_r$. In particular, letting $n_r \equiv \sum_{i=1}^n\mathbbm{1} \left \{b_i=b_r \right \}$, these distributions are:
    \begin{align*}
        \begin{split}
            &\delta_r|\pi,b,\sigma_r^2 \sim N \left ( \left [ \frac{n_r}{\sigma_r^2} + V_r \right ]^{-1} \left [ \frac{\sum_{i=1}^n\mathbbm{1} \left \{b_i=b_r \right \}\pi_i}{\sigma_r^2} + V_r\mu_r \right ] , \left [ \frac{n_r}{\sigma_r^2} + V_r \right ]^{-1} \right ) \\
            &\sigma_r^{-2}|\pi,b,\delta_r \sim \Gamma \left (\alpha_r + \frac{n_r}{2}, \beta_r + \frac{\sum_{i=1}^n\mathbbm{1} \left \{b_i=b_r \right \} \left ( \pi_i - \delta_r \right )^2}{2} \right )
        \end{split}
    \end{align*}

    

        

    \item After obtaining $Q$ draws from the posterior distributions, compute $\hat{\delta}_r=\frac{1}{Q}\sum_{q=1}^Q \delta_r^{(q)}$ for all $r$.

    \item Update bid to $\max \left \{ b_{r^*},0 \right \}$, where $r^*={\arg\max}_{r=1,\dots,R}\hat{\delta}_r$.

    \item At the end of the experiment, set $\widehat{ATE}=b_{r^*}$ if the auction is an SPA.
    
\end{enumerate}

Notice that because this approach focuses solely on bids and rewards, but not the underlying structure of the data, it is not possible to leverage Proposition \ref{prop:alignFPA} to obtain an estimate of the $ATE$ when the auction format is FPA. However, it is still possible to use Proposition \ref{prop:alignSPA} to obtain such an estimate when the auction format is SPA.

\paragraph{Correlated Thompson Sampling (TS\_CORR)} \quad There are at least two issues with TS\_OTS. First, there is a need to discretize bids, and there might be little guidance as to how to perform such discretization. Second, it ignores the information that can be obtained by exploiting the correlation in rewards across arms. This version of TS\_CORR addresses both these issues by changing the likelihood function. We now set:
\begin{enumerate}

    \item The likelihood is given by $\pi_i(b)\sim N(\delta_0 + \delta_1 b + \delta_2 b^2 + \delta_3 b^3,\sigma^2)$.
    
    \item The priors are $\delta \equiv [\delta_0,\delta_1,\delta_2,\delta_3]' \sim N(\mu,V^{-1})$ and $\sigma^{-2}\sim \Gamma(\alpha,\beta)$.
    
    \item For each $i$, after submitting bid $b_i$: 
    
    \begin{enumerate}
    
        \item Compute $\pi_i = Y_i - D_i B_{CP,i}$ if the auction is an SPA.

        \item Compute $\pi_i = Y_i - D_i b_i$ if the auction is an FPA.
        
    \end{enumerate}

    \item Draw from the posterior distribution of $\delta$ and $\sigma^2$ using all $n$ observations. Letting $\vec{b}_i \equiv [1,b_i,b_i^2,b_i^3]'$, these distributions are:

    \begin{align*}
        \begin{split}
            & \delta|\pi,b,\sigma^2 \sim N \left ( \left [ \frac{\sum_{i=1}^n \vec{b}_i \vec{b}_i'}{\sigma^2} + V \right ]^{-1} \left [ \frac{\sum_{i=1}^n \vec{b}_i \pi_i}{\sigma^2} + V \mu \right ] , \left [ \frac{\sum_{i=1}^n \vec{b}_i \vec{b}_i'}{\sigma^2} + V \right ]^{-1} \right ) \\
            &\sigma^{-2}|\pi,b,\delta \sim \Gamma \left (\alpha + \frac{n}{2}, \beta + \frac{\sum_{i=1}^n \left (\pi_i - \vec{b}_i'\delta \right )^2 }{2} \right )
        \end{split}
    \end{align*}

    
        


    \item After obtaining $Q$ draws from the posterior distributions, estimate the expected reward as a function of bids by computing $\hat{\pi}(b)=\frac{1}{Q}\sum_{q=1}^Q \vec{b}'\delta^{(q)}$.

    \item Update bid to $\max\{0,b_*\}$, where $b_* ={\arg\max}_{b\in\mathbb{R}_+}\hat{\pi}(b)$.

    \item At the end of the experiment, set $\widehat{ATE}=b_*$ if the auction is an SPA.
    
\end{enumerate}

Even though TS\_CORR arguably improves over TS\_OTS by imposing a reduced-form correlation of rewards across arms, that is, bids, which are also no longer discretized, the algorithm still disregards the underlying structure of the data, and particularly the relationship between observed and potential outcomes with treatment. Thus, for the same reason as TS\_OTS, it can obtain an estimate of $ATE$ when the auction format is SPA but not when it is FPA.

Notice that this version of TS\_CORR treats the vector $X$ as the context but, like TS\_OTS, updates the estimates of the parameters for each value of $X$ separately. We expect that this does not put these methods at an enhanced disadvantage relative to BITS in our specific setting because $X$ is a vector of mutually exclusive dummies. Given the structure from equation (\ref{eq:normal}), BITS only uses observations with different values of $X$ to learn the scale parameters of the distributions of potential outcomes and $B_{CP}$. In turn, TS\_CORR and TS\_OTS assume that rewards follow a normal distribution and only use the mean parameters for learning and optimization; these methods disregard the scale parameters. 

This specification of TS\_CORR and TS\_OTS could be more problematic if the vector $X$ was in a sense richer by, for instance, containing continuous variables. It would be straightforward to adjust these methods to contemplate such situations by specifying rewards as following a normal distribution with means equal to a polynomial of bids multiplied by some function of $X$.

\paragraph{Naive Bidding Thompson Sampling (BITS\_NAIVE)} \quad The naive implementation of BITS uses simpler distributional assumptions in its implementation. In particular, the likelihood function is now given by:
\begin{align}\label{eq:like_simple}
\begin{split}
\begin{bmatrix}\log Y_i(1) \\ \log Y_i(0) \\ \log B_{CP,i} \end{bmatrix} \Big\vert X_i &\overset{i.i.d.}{\sim} N \left ( \begin{bmatrix} X_i'\delta_{1} \\ X_i'\delta_{0} \\ X_i'\delta_{CP} \end{bmatrix}, \begin{bmatrix} \sigma_{1}^2 & \rho \sigma_{1} \sigma_{0} & 0 \\ \rho \sigma_{1} \sigma_{0} & \sigma_{0}^2 & 0 \\ 0 & 0 & \sigma_{CP}^2 \end{bmatrix}  \right ).
\end{split}
\end{align}
There are two main differences between this formulation and the main formulation of BITS. First, the potential outcomes now follow a single lognormal distribution, and not a mixture of lognormal distributions. Second, $B_{CP}$ now follows a lognormal distribution instead of a Fr\'{e}chet. As we show below, this second change leads to a much simpler algorithm to obtain draws from the posterior distribution of $\theta_{CP}$ given the data. However, it is important to note that this distributional assumption is motivated solely by convenience. Unlike the Fr\'{e}chet, it is difficult to provide a theoretical backing for this parametrization. Also, notice that by setting the correlation between $\log B_{CP}$ and $\log Y(1)$ and $\log Y(0)$ to zero, this specification imposes Assumption \ref{ass:pv}. 

For the sake convenience, we now use the following priors:
\begin{align}\label{eq:prior_simple}
    \begin{split}
        &\delta_j\sim N(\mu_j,V_j^{-1}) \\
        &\sigma_j^{-2}\sim \Gamma(\alpha_j,\beta_j)^2
    \end{split}
\end{align}
where $j\in\{1,0,CP\}$. Thus, this implementation also treats $Y(1)$ and $Y(0)$ as independent even though they might not be.

The procedure to obtain draws from the posterior distributions of $\theta_1$ and $\theta_0$ coincides with the main procedure with $K=1$. The procedure to obtain draws from the posterior distribution of $\theta_{CP}$ given the data, however, is simpler, because now we can also use Gibbs sampling with data augmentation instead of Metropolis-Hastings. 

To see this, notice that (\ref{eq:prior_simple}) implies that $\frac{\log B_{CP}-X_i'\delta_{CP}}{\sigma_{CP}}\sim N(0,1)$. Consequently, normalizing $\sigma_{CP}=1$, it follows that $D=\mathbbm{1}\left \{ \log b - X_i'\delta_{CP} + \epsilon_{CP} \geq 0 \right \}$, where $\epsilon_{CP}\sim N(0,1)$. This corresponds to a Probit model, so that we can follow the standard Bayesian method, which uses Gibbs sampling with data augmentation, to obtain draws from the posterior distribution of $\delta_{CP}$ given the data.

We can summarize the BITS\_NAIVE procedure as follows:
\begin{enumerate}

    \item The likelihood is given in (\ref{eq:like_simple}).
    
    \item The priors are given in (\ref{eq:prior_simple}).
    
    \item Obtain $Q$ draws from the posterior distributions of $\theta_1$ and $\theta_0$ following the procedure from Appendix \ref{app:potout} and compute:
    \begin{align}\label{eq:ate_simple}
        \widehat{ATE}(x)=\frac{1}{Q} \sum_{q=1}^Q \left ( \exp \left \{x'\delta_1^{(q)} + 0.5 \sigma_1^{2,(q)} \right \} - \exp \left \{x'\delta_0^{(q)} + 0.5 \sigma_0^{2,(q)} \right \} \right ).   
    \end{align} 

    \item If the auction is an FPA, obtain $Q$ draws from the posterior distribution of $\delta_{CP}$ using the procedure from \cite{ac1993} and compute:
    \begin{align}\label{eq:chi_simple}
        \hat{\chi}(b|x)= b  \left [ 1 + \frac{1}{Q} \sum_{q=1}^Q \frac{\Phi \left (\log b - x'\delta_{CP}^{(q)} \right )}{\phi \left (\log b - x'\delta_{CP}^{(q)} \right )} \right ],
    \end{align}
    where $\phi(\cdot)$ and $\Phi(\cdot)$ denote the pdf and cdf of the standard normal distribution, respectively. 

    \item If the auction is an SPA, update bid to $\max \left \{ 0, \widehat{ATE}(x) \right \}$. If the auction is an FPA, update the bid to $\max \left \{ 0, \hat{\chi}^{-1} \left [\widehat{ATE}(x) | x \right ]  \right \}$, where $\widehat{ATE}(x)$ is given in (\ref{eq:ate_simple}) and $\hat{\chi}(\cdot|x)$ is given in (\ref{eq:chi_simple}). Notice that the standard normal distribution has a decreasing reversed hazard rate, so $\hat{\chi}(\cdot|x)$ is strictly increasing and therefore its inverse is well-defined.

    \item At the end of the experiment, the estimate of $ATE(x)$ is obtained from (\ref{eq:ate_simple}). 

\end{enumerate}

As the main specification of BITS, BITS\_NAIVE exploits the structure of the problem to account for the correlation of rewards across arms and contexts, and thus it can leverage both Propositions \ref{prop:alignSPA} and \ref{prop:alignFPA}. Furthermore, the algorithms used to implement BITS\_NAIVE are simpler and faster. However, this convenience comes with a decrease in the flexibility of the model.

\paragraph{On greedy algorithms} \quad It is straightforward to implement greedy frequentist counterparts to the four TS-based algorithms we consider in our simulations. This approach would be based on maximum likelihood instead of on the posterior distribution of parameters given the data. 

A greedy algorithm can be implemented by: (a) posing some parametric model, such as the ones we consider above; (b) using the data to estimate the parameters of these models via maximum likelihood; (c) computing the optimal bids based on these point estimates by, for example, using Propositions \ref{prop:alignSPA} and \ref{prop:alignFPA} or equation; and (d) submitting these estimated optimal bids. 

For practical purposes, these approaches might be preferable in situations where maximizing the likelihood function is easier than obtaining draws from the posterior distribution. However, these greedy frequentist methods do not explicitly account for the serial correlation in the data. Hence, they do not output objects that allows one to perform inference on the estimated treatment effect parameters.

\subsection{Full set of results}\label{app:full_res}

\begin{figure}[h]
    \centering
	\includegraphics[width=.8\textwidth]{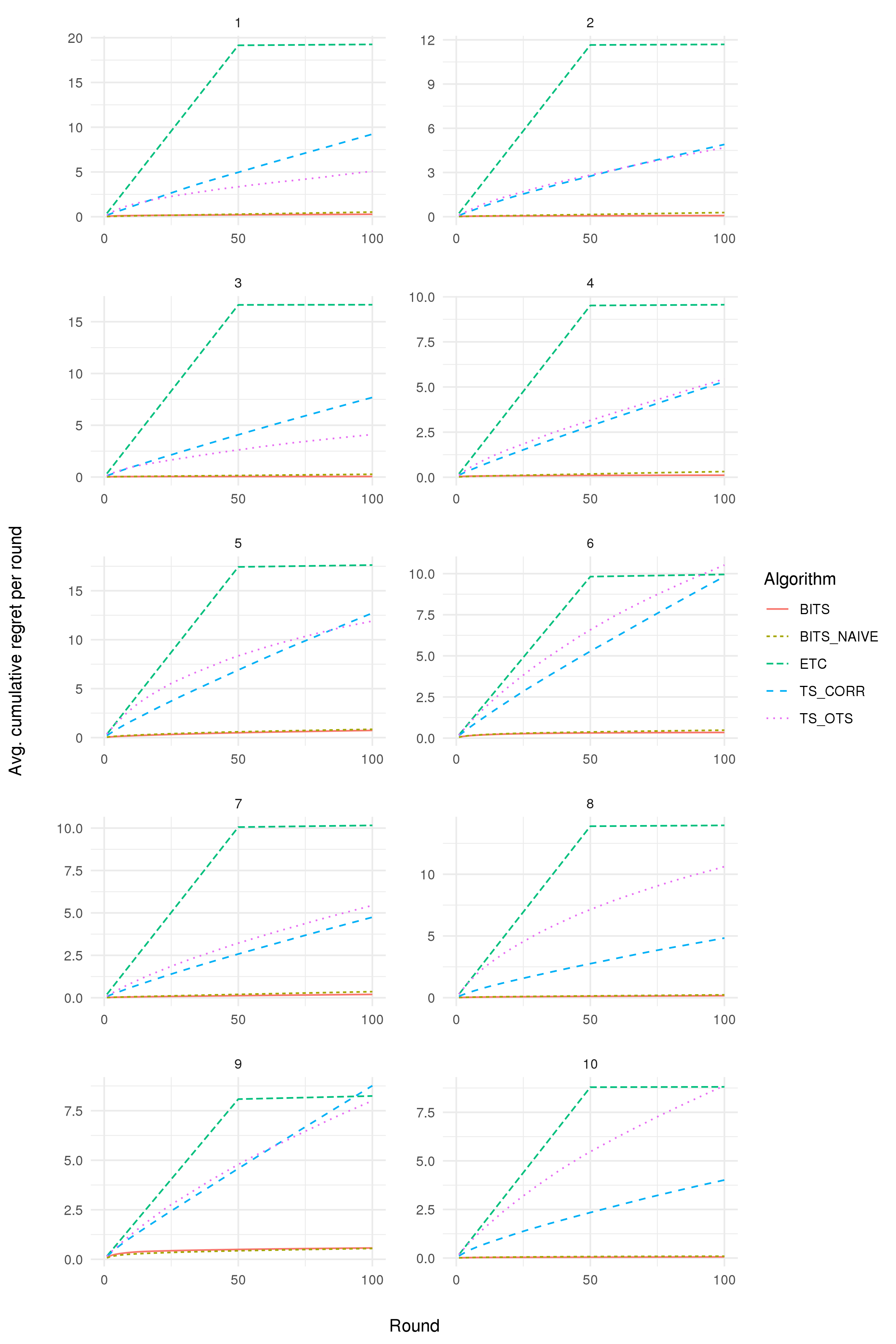}
    \caption{Evolution of cumulative regret for SPAs}
    \label{fig:reg_SPA}%
\end{figure}

\begin{figure}[h]
    \centering
	\includegraphics[width=.8\textwidth]{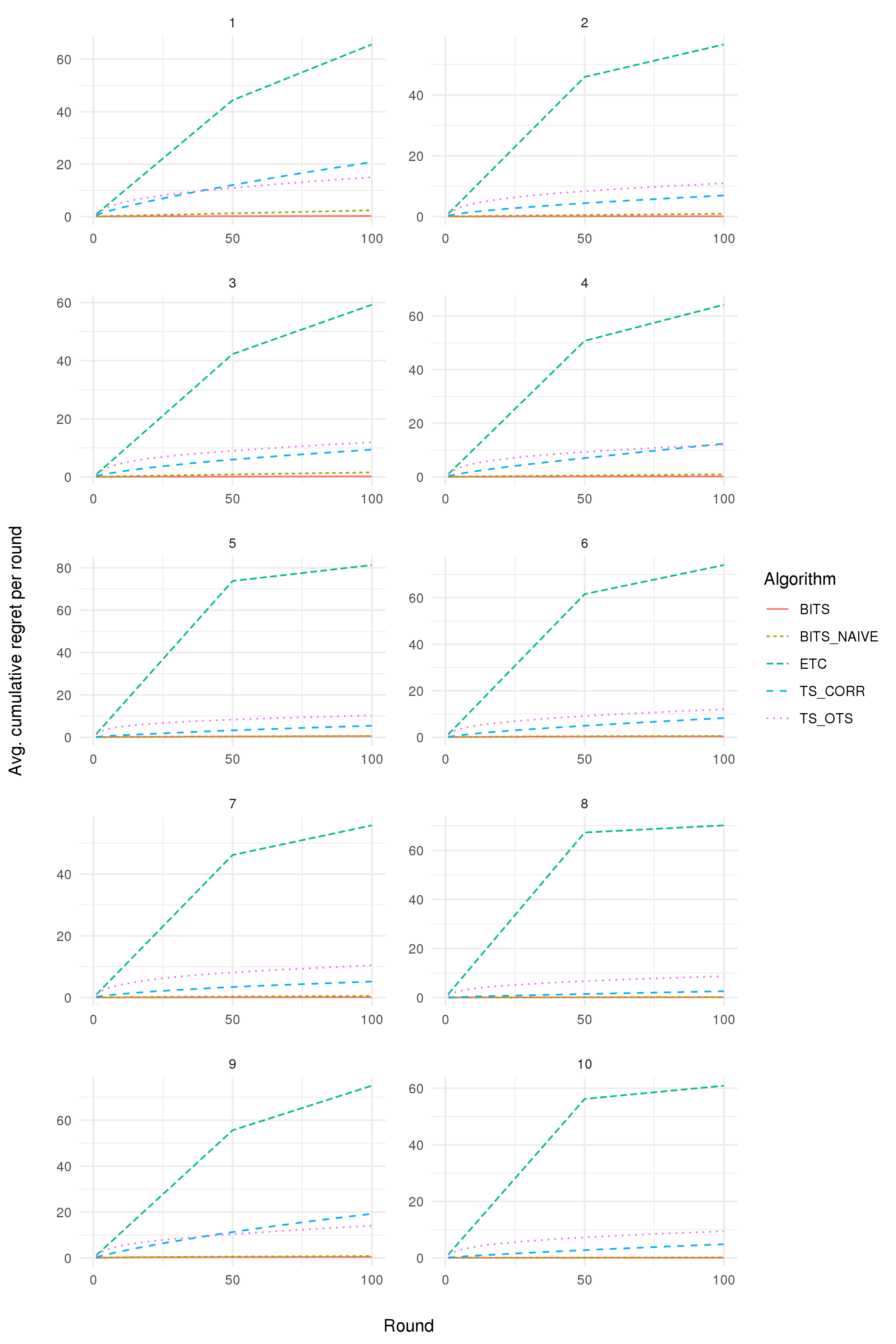}
    \caption{Evolution of cumulative regret for FPAs}
    \label{fig:reg_FPA}%
\end{figure}

Figures \ref{fig:reg_SPA} and \ref{fig:reg_FPA} are analogous to Figures \ref{fig:reg_ctxt_spa} and \ref{fig:reg_ctxt_fpa}, respectively, and show the evolution of cumulative regret averaged across the $E$ epochs for all contexts considered in the simulations. Overall, these results are qualitatively equivalent.

\begin{sidewaysfigure}[h]
    \centering
    \begin{subfloat}[\scriptsize Context 1\label{fig:mse_spa1}]
        {\includegraphics[page=1, width=0.15\textwidth]{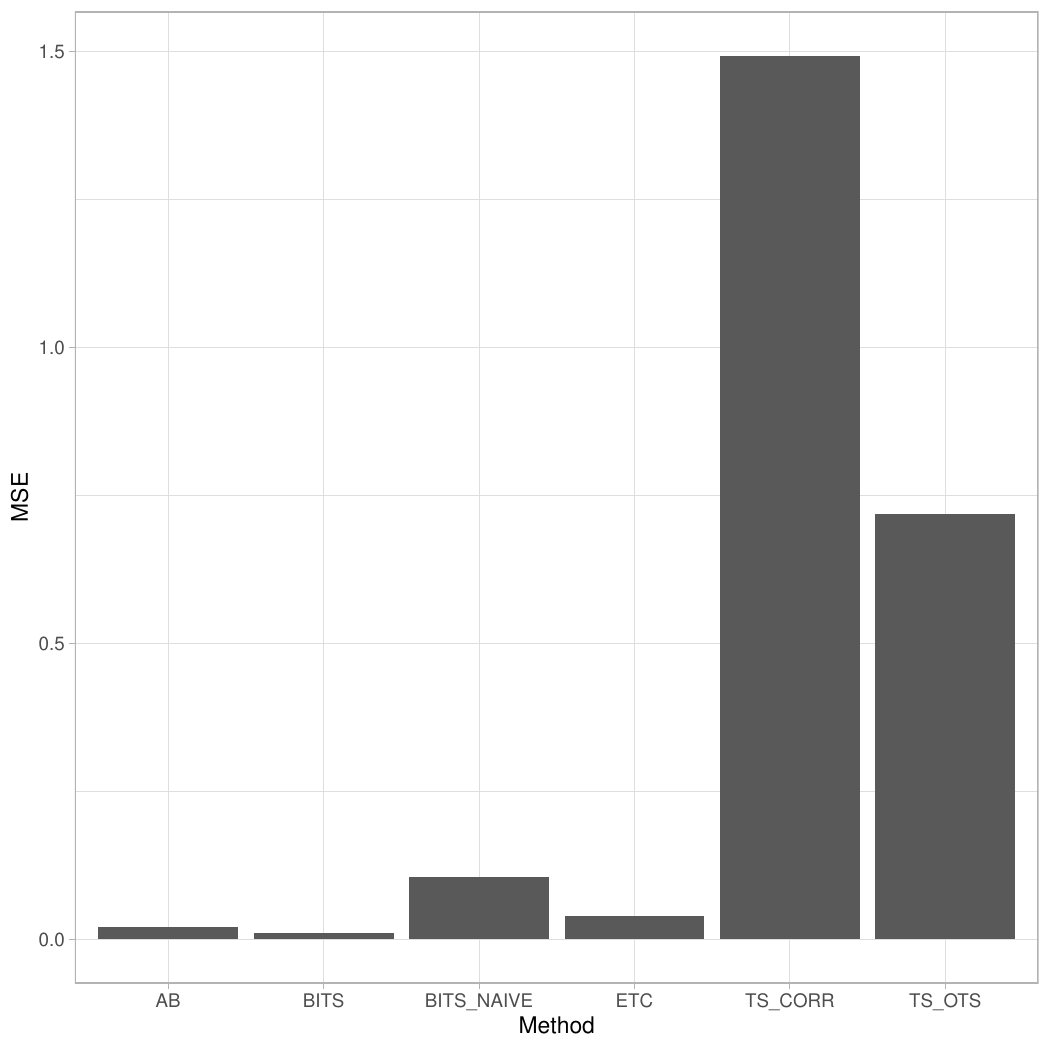}}
    \end{subfloat}
    \begin{subfloat}[\scriptsize Context 2\label{fig:mse_spa2}]
        {\includegraphics[page=2, width=0.15\textwidth]{mse_spa.pdf}}
    \end{subfloat}
    \begin{subfloat}[\scriptsize Context 3\label{fig:mse_spa3}]
        {\includegraphics[page=3, width=0.15\textwidth]{mse_spa.pdf}}
    \end{subfloat}
    \begin{subfloat}[\scriptsize Context 4\label{fig:mse_spa4}]
        {\includegraphics[page=4, width=0.15\textwidth]{mse_spa.pdf}}
    \end{subfloat}
    \begin{subfloat}[\scriptsize Context 5\label{fig:mse_spa5}]
        {\includegraphics[page=5, width=0.15\textwidth]{mse_spa.pdf}}
    \end{subfloat}
    \\
    \begin{subfloat}[\scriptsize Context 1\label{fig:mse_spa6}]
        {\includegraphics[page=6, width=0.15\textwidth]{mse_spa.pdf}}
    \end{subfloat}
    \begin{subfloat}[\scriptsize Context 2\label{fig:mse_spa7}]
        {\includegraphics[page=7, width=0.15\textwidth]{mse_spa.pdf}}
    \end{subfloat}
    \begin{subfloat}[\scriptsize Context 3\label{fig:mse_spa8}]
        {\includegraphics[page=8, width=0.15\textwidth]{mse_spa.pdf}}
    \end{subfloat}
    \begin{subfloat}[\scriptsize Context 4\label{fig:mse_spa9}]
        {\includegraphics[page=9, width=0.15\textwidth]{mse_spa.pdf}}
    \end{subfloat}
    \begin{subfloat}[\scriptsize Context 5\label{fig:mse_spa10}]
        {\includegraphics[page=10, width=0.15\textwidth]{mse_spa.pdf}}
    \end{subfloat}
\caption{MSEs for SPAs}
    \label{fig:mse_spa}%
\end{sidewaysfigure}

\begin{sidewaysfigure}[h]
    \centering
    \begin{subfloat}[\scriptsize Context 1\label{fig:mse_fpa1}]
        {\includegraphics[page=1, width=0.15\textwidth]{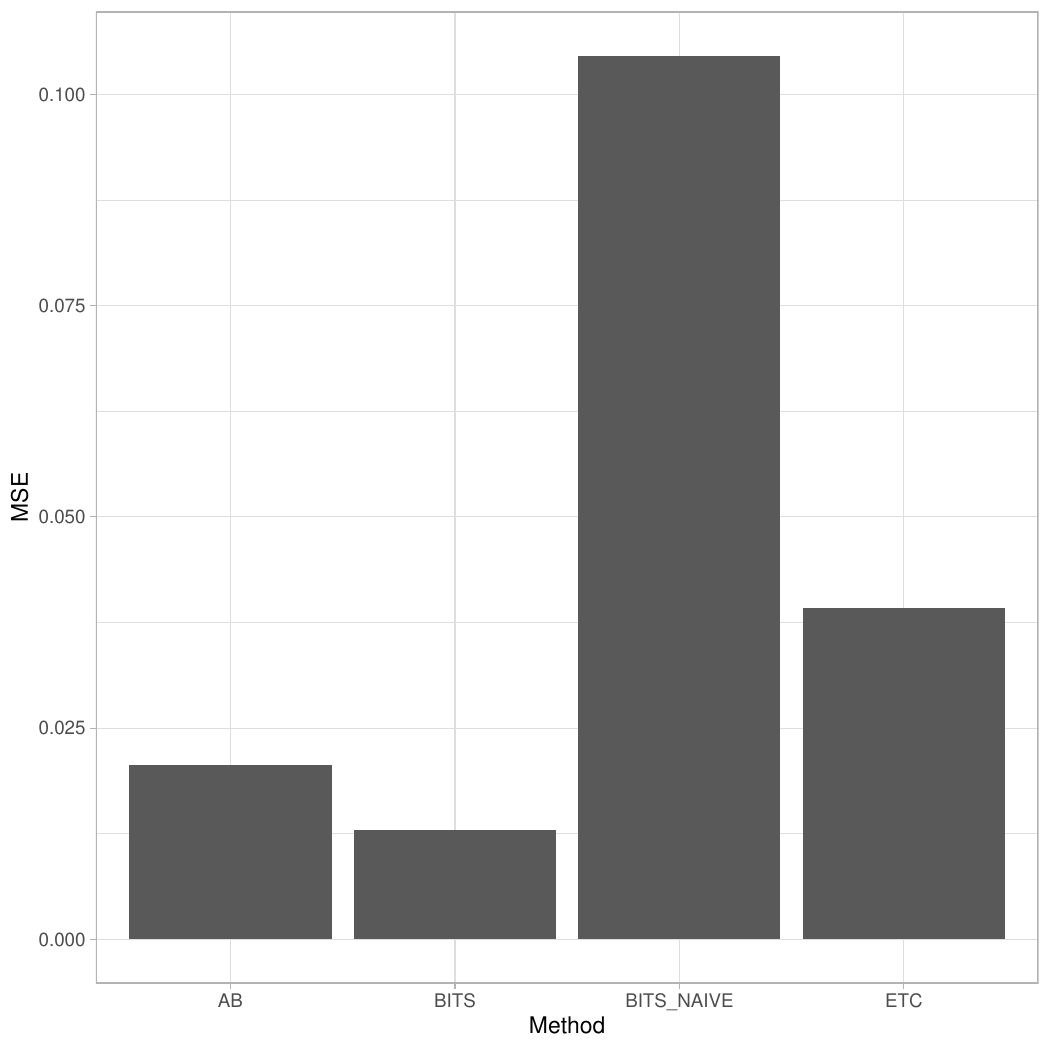}}
    \end{subfloat}
    \begin{subfloat}[\scriptsize Context 2\label{fig:mse_fpa2}]
        {\includegraphics[page=2, width=0.15\textwidth]{mse_fpa.pdf}}
    \end{subfloat}
    \begin{subfloat}[\scriptsize Context 3\label{fig:mse_fpa3}]
        {\includegraphics[page=3, width=0.15\textwidth]{mse_fpa.pdf}}
    \end{subfloat}
    \begin{subfloat}[\scriptsize Context 4\label{fig:mse_fpa4}]
        {\includegraphics[page=4, width=0.15\textwidth]{mse_fpa.pdf}}
    \end{subfloat}
    \begin{subfloat}[\scriptsize Context 5\label{fig:mse_fpa5}]
        {\includegraphics[page=5, width=0.15\textwidth]{mse_fpa.pdf}}
    \end{subfloat}
    \\
    \begin{subfloat}[\scriptsize Context 1\label{fig:mse_fpa6}]
        {\includegraphics[page=6, width=0.15\textwidth]{mse_fpa.pdf}}
    \end{subfloat}
    \begin{subfloat}[\scriptsize Context 2\label{fig:mse_fpa7}]
        {\includegraphics[page=7, width=0.15\textwidth]{mse_fpa.pdf}}
    \end{subfloat}
    \begin{subfloat}[\scriptsize Context 3\label{fig:mse_fpa8}]
        {\includegraphics[page=8, width=0.15\textwidth]{mse_fpa.pdf}}
    \end{subfloat}
    \begin{subfloat}[\scriptsize Context 4\label{fig:mse_fpa9}]
        {\includegraphics[page=9, width=0.15\textwidth]{mse_fpa.pdf}}
    \end{subfloat}
    \begin{subfloat}[\scriptsize Context 5\label{fig:mse_fpa10}]
        {\includegraphics[page=10, width=0.15\textwidth]{mse_fpa.pdf}}
    \end{subfloat}
\caption{MSEs for FPAs}
    \label{fig:mse_fpa}%
\end{sidewaysfigure}

In turn, Figures \ref{fig:mse_spa} and \ref{fig:mse_fpa} are analogous to Figures \ref{fig:mse_ctxt_spa} and \ref{fig:mse_ctxt_fpa}, respectively, and show the MSEs of the methods under consideration for all contexts. Once again, the results are qualitatively equivalent.

\end{document}